\documentclass[sn-mathphys-num,iicol,preprint]{sn-jnl}

\pdfoutput=1
\usepackage[utf8]{inputenc} 
\usepackage[T1]{fontenc}    
\usepackage{hyperref}       
\usepackage{nicefrac}       
\usepackage{microtype}      
\usepackage{colortbl}
\usepackage{xspace}
\usepackage{tikz}
\usepackage{caption}
\usepackage{adjustbox}
\usepackage{rotating}
\usepackage{soul}
\usepackage{subcaption}
\usepackage{sidecap}
\usepackage{graphicx}%
\usepackage{multirow}%
\usepackage{amsmath,amssymb,amsfonts}%
\usepackage{amsthm}%
\usepackage{mathrsfs}%
\usepackage[title]{appendix}%
\usepackage{xcolor}%
\usepackage{textcomp}%
\usepackage{manyfoot}%
\usepackage{booktabs}%
\usepackage{algorithm}%
\usepackage{algorithmicx}%
\usepackage{algpseudocode}%
\usepackage{listings}%
\usepackage{tabularx}
\usepackage{array}
\usepackage{nicematrix}
\usepackage{tabularray}
\usepackage{lmodern}
\usepackage{xcolor,pifont}
\usepackage{float}
\usepackage{stfloats}
\usepackage{afterpage}
\newcommand{\cmark}{\ding{51}} 
\newcommand{\xmark}{\ding{55}} 

\newcommand{\greencheck}{{\color{green}\cmark}}
\newcommand{\redx}{{\color{red}\xmark}}

\newif\ifcomments

\commentsfalse

\ifcomments
\newcommand{\comments}[1]{#1}
\else
\newcommand{\comments}[1]{}
\fi

\newcommand{\andrija}[1]{\comments{\textcolor{blue}{[andrija: #1]}}}
\newcommand{\mladen}[1]{\comments{\textcolor{olive}{[mladen: #1]}}}
\newcommand{\rosanne}[1]{\comments{\textcolor{red}{[rosanne: #1]}}}

\newcommand{\todo}[1]{\comments{\textcolor{red}{[TODO: #1]}}}

\definecolor{muchlater}{rgb}{0.7,1,.7}

\newcolumntype{M}[1]{>{\raggedright\arraybackslash}m{#1cm}}
\newcolumntype{N}[1]{>{\centering\arraybackslash}m{#1cm}}

\newcommand{\figlabel}[1]{\label{fig:#1}}
\newcommand{\figref}[1]{Figure~\ref{fig:#1}}

\newcommand{\tablabel}[1]{\label{tab:#1}}
\newcommand{\tabref}[1]{Table~\ref{tab:#1}}

\newcommand{\alglabel}[1]{\label{alg:#1}}
\newcommand{\algoref}[1]{Algorithm~\ref{alg:#1}}

\definecolor{ltscolor}{rgb}{0.7, 0.35, 0.35}
\definecolor{lightgray}{gray}{0.9}

\begin{document}

\title{Logit Scaling for Out-of-Distribution Detection}

\author*[1,2]{\fnm{Andrija} \sur{Djurisic}}\email{andrija@mlcollective.org}
\author[2,3]{\fnm{Rosanne} \sur{Liu}}
\author[1]{\fnm{Mladen} \sur{Nikolic}}

\affil[1]{\orgdiv{Faculty of Mathematics}, \orgname{University of Belgrade}, \country{Serbia}}
\affil[2]{\orgname{ML Collective}}
\affil[3]{\orgname{Google DeepMind}}

\abstract{
The safe deployment of machine learning and AI models in open-world settings hinges critically on the ability to detect out-of-distribution (OOD) data accurately, data samples that contrast vastly from what the model was trained with. 
Current approaches to OOD detection often require further training the model, and/or statistics about the training data which may no longer be accessible. Additionally, many existing OOD detection methods struggle to maintain performance when transferred across different architectures. Our research tackles these issues by proposing a simple, post-hoc method that does not require access to the training data distribution, keeps a trained network intact, and holds strong performance across a variety of architectures. 
Our method, Logit Scaling (\textbf{LTS}), as the name suggests, simply scales the logits in a manner that effectively distinguishes between in-distribution (ID) and OOD samples. We tested our method on benchmarks across various scales, including CIFAR-10, CIFAR-100, ImageNet and OpenOOD. The experiments cover 3 ID and 14 OOD datasets, as well as 9 model architectures. Overall, we demonstrate state-of-the-art performance, robustness and adaptability across different architectures, paving the way towards a universally applicable solution for advanced OOD detection. Our code is available at \url{http://github.com/andrijazz/lts}.

}

\keywords{out-of-distribution detection, distribution shift, robustness, post hoc, generalization}

\maketitle

\section{Introduction}

\begin{figure*}[!t]
    \centering 
	\includegraphics[width=\linewidth]{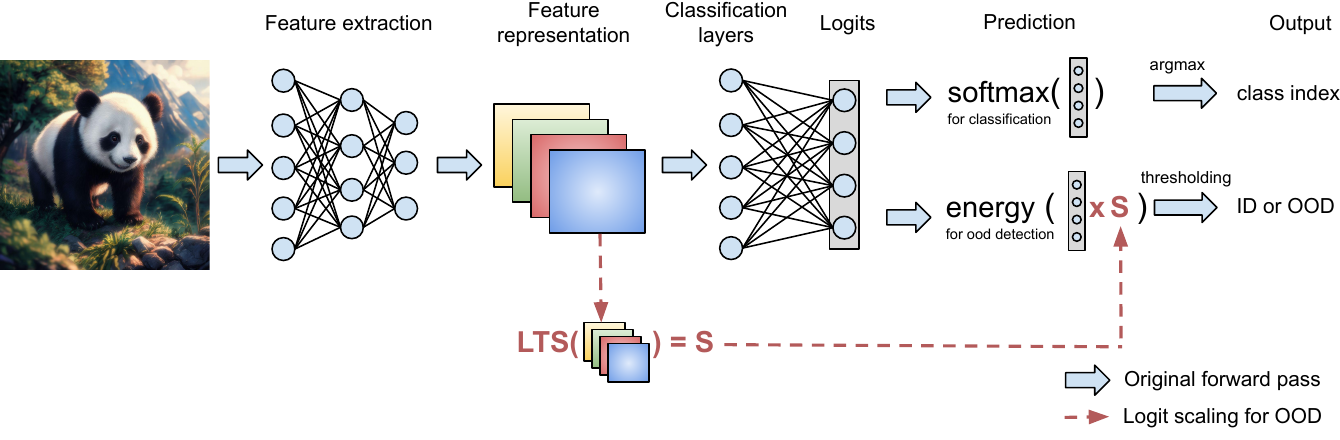}
	\caption{\textbf{Overview of the LTS method for OOD detection.} LTS works at inference time during forward pass. It takes features representations and computes sample-specific scalar value which is then used to scale the logits. Final OOD detection score is calculated by applying scoring function to scaled logits. LTS incurs minimal computational costs and it doesn't modify activations in any way thus completely preserves original working of the network while enhancing OOD detection significantly.} 
	\figlabel{fig:overview}
\end{figure*}

In the forthcoming era of artificial intelligence (AI), ensuring the safety and reliability of AI models is of utmost importance~\citep{amodei2016concrete, mohseni2106practical, hendrycks2021unsolved}. One of the key components in achieving this is the effective detection of out-of-distribution (OOD) data on which the trained models usually severely underperform~\citep{Nguyen2015, hendrycks17baseline}. Such underperformance poses a significant challenge to model reliability and safety. This is particularly crucial in areas such as healthcare, autonomous driving, and finance where the ability to detect OOD instances can profoundly influence outcomes. Thus, ensuring models can accurately recognize and manage OOD samples not only enhances their performance but also plays a vital role in preventing potential errors in critical applications.

The flip side of OOD is the in-distribution (ID) data, which usually correspond to the exact training data of the model, or data samples from the same distribution as the training data. AI and machine learning research preceding the current phase of large models rely on standardized training sets that are made open and accessible, but the situation has changed starting from the large models era, where the majority of state-of-the-art models have their training data undisclosed or difficult to access. In light of this change, we are in greater need of OOD detection methods that do not necessarily require full access to the model's training data.

The line of work in OOD detection has advanced drastically in recent years. ReAct~\citep{react} first revealed that the activation patterns in the penultimate layer of the networks conceal valuable information for identifying out-of-distribution samples. ReAct~\citep{react} achieved strong OOD performance by clipping activations of the penultimate layer utilizing training data statistics. Furthermore, the methods ASH~\citep{djurisic2022extremely} and SCALE~\citep{xu2023scaling} demonstrated that adjusting activations in the penultimate layer of the network by pruning and scaling on a per-sample basis leads to strong OOD detection performance without access to in-distribution data. Both methods delivered robust OOD performance, yet altering activations also resulted in a slight reduction in model accuracy. Moreover, neither of said methods demonstrated successful OOD detection over a diverse set of neural network architectures. 

\newcolumntype{F}[1]{%
    >{\raggedright\arraybackslash\hspace{0pt}}p{#1}}%
\newcolumntype{T}[1]{%
    >{\centering\arraybackslash\hspace{0pt}}p{#1}}%

\begin{table}[!h]
\centering
\resizebox{\linewidth}{!}{
\begin{tabular}
{T{0.3\linewidth}|T{0.2\linewidth}|T{0.2\linewidth}|T{0.2\linewidth} T{0.2\linewidth}}
\hline
\textbf{Method} & \textbf{Works without training data access} & \textbf{Doesn't require OOD preparation steps} & \textbf{Doesn't affect ID accuracy} \\
\hline
REACT~\citep{react} & \redx & \redx & \redx  \\
DICE~\citep{sun2022dice} & \redx & \redx & \redx  \\
ASH~\citep{djurisic2022extremely} & \greencheck & \greencheck & \redx  \\
SCALE~\citep{xu2023scaling} & \greencheck & \greencheck & \redx  \\
OptFS~\citep{zhao2024towards} & \redx & \redx & \greencheck \\
ATS~\citep{krumpl2024ats} & \redx & \redx & \greencheck \\
\rowcolor{lightgray}\textbf{LTS (Ours)} & \greencheck & \greencheck & \greencheck  \\
\hline
\end{tabular}
}
\caption{\textbf{Preferable properties of a OOD detection methods}. Table compares key properties of various OOD detection methods, highlighting whether each method operates without training data access, avoids the need for OOD-specific preparation steps (which usually involve deriving statistics from training data), and maintains in-distribution accuracy. Our proposed method, LTS, satisfies all three criteria.}
\tablabel{checkbox_table}
\end{table}

In this paper, we present a \textbf{l}ogi\textbf{t} \textbf{s}caling (LTS) method for out-of-distribution detection. Our method operates completely post-hoc, meaning that model does not rely on the statistics derived from the training data. This method is ready to use off-the-shelf, offering a remarkably simple, flexible and cost-effective solution. LTS leverages insights from \citet{react, djurisic2022extremely, xu2023scaling} and relies on the feature representation from the penultimate layer. Based on that respresentation, it computes a scaling factor on a per-sample basis by using the relationship between strong and week activations. The scaling factor is used to scale the logits. Once scaled, logits are then used to compute the final OOD score using well established OOD scoring function, {\em energy score}~\citep{liu2020energy}. \figref{fig:overview} summarizes the inner workings of our method.

Numerous previous methods for OOD detection were constrained to a limited set of architectures. Recent research by~\citet{zhao2024towards} conducted a thorough analysis of this issue, revealing that while many methods offer robust OOD performance for some architectures, they lack applicability across different architectures. Their method, OptFS~\citep{zhao2024towards}, demonstrates remarkable results on architectures for which most previous methods struggled. 
We evaluated LTS across 9 distinct architectures and achieved superior performance compared to previous state-of-the art method, OptFS~\citep{zhao2024towards}.

\tabref{checkbox_table} presents an overview of the desirable properties of OOD detection methods and compares all the OOD detection methods discussed.

Our work makes the following contributions to the field of OOD detection:
\begin{itemize}
    \item We propose a simple and post-hoc method LTS, that can be used off-the-shelf without access to in-distribution data. \rosanne{As simple as ASH}
    \item We demonstrate that the LTS achieves new state-of-the-art results over 3 in-distribution (ID) and 14 out-of-distribution (OOD) datasets. \rosanne{Lots of exps; SOTA}
    \item We demonstrate that LTS is showing robust OOD detection performance across nine different architectures, significantly reducing FPR@95 while maintaining AUROC compared to previous state-of-the-art method. \rosanne{Transfer across architectures}
\end{itemize}

\section{Related work}

The field of out-of-distribution detection has expanded significantly in recent years, driven by the need for safer ML model deployment. Baseline methods such as those proposed by \citet{hendrycks17baseline}, \citet{liu2020energy}, and \citet{mahalanobis} have laid the foundation for much of the subsequent research in this area. We discuss four major groups of OOD detection methods relevant for understanding our own work.

\textbf{Methods relying on training data statistics.} Techniques such as ReAct~\citep{react}, DICE~\citep{sun2022dice}, KNN-based methods~\citep{sun2022out} and ATS~\citep{krumpl2024ats} use in-distribution data to derive thresholds or calibration parameters. These methods are effective but require access to training data and often involve additional processing steps after training. 

\textbf{Post-hoc model enhancement methods.} To address the impracticality of accessing large-scale training data, methods such as ASH~\citep{djurisic2022extremely} and SCALE~\citep{xu2023scaling} modify penultimate-layer activations per sample without statistical information derived from training data. Although these approaches improve detection, they can negatively affect the network’s performance on the original task. LTS similarly operates post hoc but avoids modifying activations directly, thereby completely preserving ID performance and leaving network inner working intact. Additional details on the differences between our method and SCALE are provided in Appendix~\ref{appendix_diff_lts_vs_scale}.

\textbf{Logit scaling and temperature-based methods.} Another family of post-processing approaches works by calibrating the output logits or softmax scores to better distinguish OOD uncertainty. Techniques such as ODIN~\citep{Liang2017} apply a fixed high-temperature scaling to soften the softmax distribution (often combined with input perturbation), while more recent methods like Adaptive Temperature Scaling (ATS)~\citep{krumpl2024ats} choose a temperature per sample based on intermediate-layer signals and in combination with empirical cumulative distribution function derived from training data statistics. Further details on the differences between our approach and ATS are provided in Appendix~\ref{appendix_diff_lts_vs_ats}.

\textbf{Training time OOD enhancement.} Recently, numerous approaches have been incorporating adjustments during the training phase to enhance OOD detection during inference~\citep{xu2023scaling, wei2022mitigating, pinto2022using}.  These approaches typically involve either architectural modifications~\citep{wei2022mitigating} or data augmentations during training~\citep{pinto2022using}. A recent method, ISH~\citep{xu2023scaling}, modifies the training process by introducing the scaling of activations during the backward pass. A major disadvantage of enhancements made during the training phase is the increased computational cost.

\section{Logit Scaling for OOD Detection}

The energy score, as introduced by~\citet{liu2020energy}, is one of the most frequently employed techniques for detecting out-of-distribution samples. The process of detecting OOD samples using energy score operates in the following manner: raw inputs are processed by the network, the logits are computed, and fed into the energy score function. This function then produces a score used to classify whether a sample is in-distribution (ID) or out-of-distribution (OOD). In this work, we propose a modification to this standard procedure.

The energy score is defined as a scalar value computed by the formula: 
$$E(\mathbf{x}; f) = - \log \sum_{i=1}^C e^{f_i(\mathbf{x})}$$ 
where $\mathbf{x}$ is the given input, $C$ is the number of classes and $f_i(\mathbf{x})$ is the logit for the class $i$. 

Our approach modifies the aforementioned process by extending it with the computation of a scaling factor $S(\mathbf{x})$ derived from the feature representation of an individual sample $\mathbf{x}$:
$$E(\mathbf{x}; f) = - \log \sum_{i=1}^C e^{S(\mathbf{x})f_i(\mathbf{x})}$$ 
The scaling factor $S$ is computed in the following manner. We denote the feature representation of an input sample $\mathbf{x}$ from the network's penultimate layer as $h(\mathbf{x}) \in \mathbb{R}^m$. Our method calculates the scaling factor $S$ by dividing the sum of all elements in $h(\mathbf{x})$ by the sum of the top $p$-th percent of $h(\mathbf{x})$ activations. \algoref{algo} outlines the internal mechanics of the LTS method. In the algorithm $f_p(\cdot)$ denotes the network up to and including the penultimate layer and $\text{logits}(\cdot)$ denotes the final layer, so the full network $f(\cdot)$ can be represented as a composition $\text{logits}\circ f_p$, meaning $f(\mathbf{x})=\text{logits}(f_p(\mathbf{x}))$ for all inputs $\mathbf{x}$. The function $\text{top}_k(h,k)$ 
returns the largest $k$ elements of $h$.

Such simple treatment increases the scaling factor $S$ for ID samples and reduces it for OOD samples, effectively influencing separation of two distributions. The sample-based scaling factor $S$ is then used to scale the logits, facilitating the use of the energy score for the final OOD detection as shown in ~\figref{fig:overview}. \figref{effect_of_lts} illustrates the impact of applying LTS on the logits distribution and the resulting OOD detection scores. 
A more detailed illustration of the effects of LTS as the hyperparameter $p$ is varied is provided in Appendix~\ref{appendix_lts_intuition}.

\begin{figure*}[h]
 \centering
 \begin{subfigure}[h]{0.49\textwidth}
     \centering
     \includegraphics[width=\textwidth]{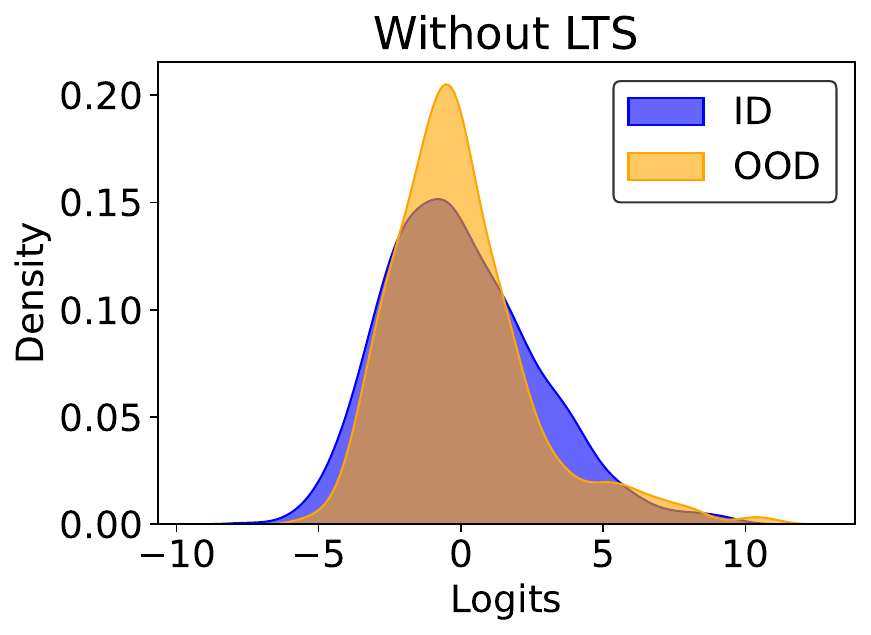}
     \label{a}
 \end{subfigure}
  \hfill
  \begin{subfigure}[h]{0.49\textwidth}
     \centering
     \includegraphics[width=\textwidth]{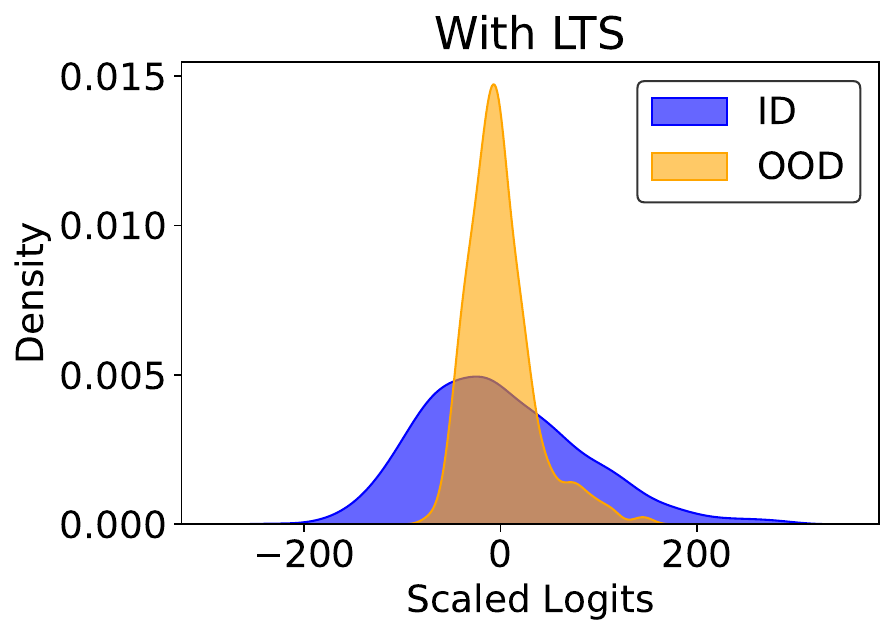}
     \label{a}
 \end{subfigure}
 \begin{subfigure}[h]{0.49\textwidth}
     \centering
     \includegraphics[width=\textwidth]{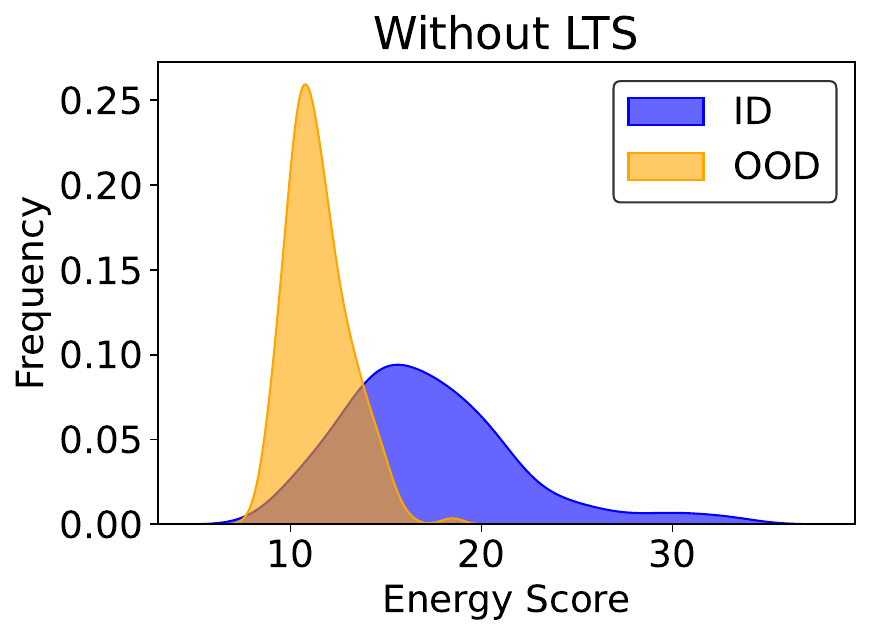}
     \label{a}
 \end{subfigure}
 \hfill
 \begin{subfigure}[h]{0.49\textwidth}
     \centering
     \includegraphics[width=\textwidth]{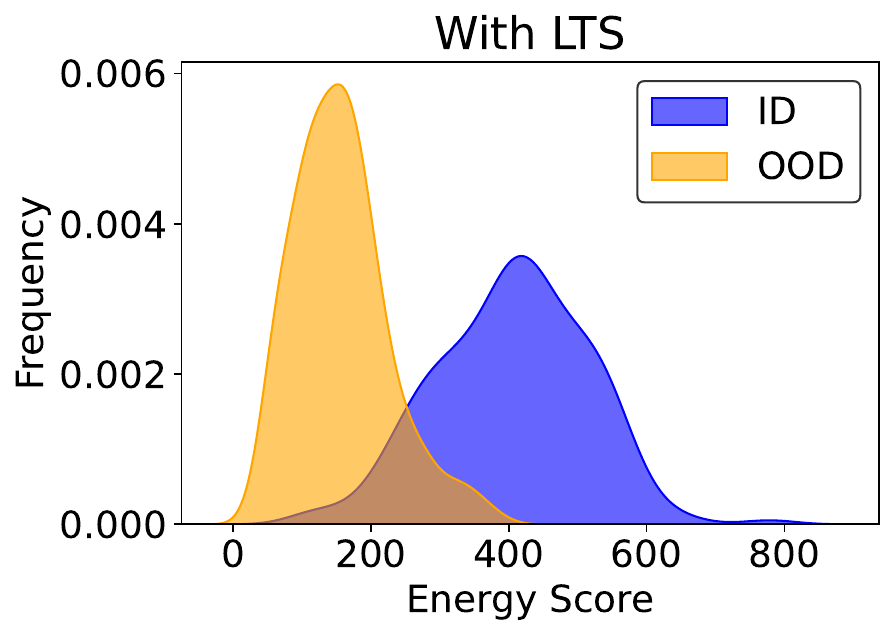}
     \label{b}
 \end{subfigure}
\caption{\textbf{Effect of LTS Treatment.} Plots demonstrate the changes in the distribution of logits and Energy scores resulting from LTS treatment. The left-hand plots represent the state before LTS application, while the right-hand plots (zoomed-in) represent the state after LTS is applied. The plots were generated using a ResNet-50 architecture pretrained on the ImageNet-1k (ID) dataset, with iNaturalist serving as the OOD dataset. The application of LTS produces more extreme logit distribution for OOD samples and improves the separation between ID and OOD scores, leading to a substantial enhancement in OOD detection performance. The logit distribution plots were generated using a single ID and a single OOD sample, whereas the energy score plots were created using 200 images sampled from both the ID and OOD datasets.}
\figlabel{effect_of_lts}
\end{figure*}

The proposed method has several desirable properties. After the forward pass is performed (which is necessary for general inference), it incurs minimal computational cost -- at its core is a simple computation over the penultimate layer. Also, it is memory-efficient, requiring just a few auxiliary variables. It does not modify the activations of the network, thus preserving the network's performance without any negative impact. It is universaly applicable, namely, it can be applied to any architecture with minimal programming effort. It performs detection based strictly on the analyzed sample and does not require the access to the data used to train the network.

\begin{algorithm}[h]
    \caption{\textbf{\textcolor{ltscolor}{Logit Scaling for OOD detection (LTS)}}}
    \alglabel{algo}
    \textbf{Input:} Single input sample $\mathbf{x}$, $\mathbf{p}$ - fraction of top activations we are considering\\
    \textbf{Output:} Scaling factor S\\
    \begin{algorithmic}[1]
        \State $h\leftarrow f_p(\mathbf{x})$
        \State $n\leftarrow \text{dim}(h)$
        \State $S1\leftarrow \sum_{i=1}^{n}h_i$
        \State $h_{top}=\text{top}_k(h,p\cdot n)$
        \State $S2\leftarrow \sum_{t\in h_{top}} t$ 
        \State $S\leftarrow \left(\frac{S1}{S2}\right)^2$ 
        \\
        \Return $S$
    \end{algorithmic}
\end{algorithm}

Our method is inspired by findings reported in ReAct~\citep{react}, which highlight that out-of-distribution samples often induce abnormally high activations. These findings are illustrated by \figref{intuition}. Moreover, in our prior work ASH~\citep{djurisic2022extremely} we observed that high-magnitude activations tend to capture the information most useful for the network’s task. Namely, we showed that it is possible to prune up to 90\% of the lowest activations while still maintaining strong model performance. These two observations justify our treatment of high-magnitude activations as particularly important for the out-of-distribution detection task. Therefore, in the LTS we define the scaling factor so as to emphasize the contribution of the strongest activations (the top $p$ percent of $h$) that are capturing the differences in activation patterns between ID and OOD data. Leveraging these discrepancies, we use the scaling factor to adjust the logits, ultimately enhancing the separation between in-distribution and out-of-distribution samples.

\begin{figure*}[h]
 \centering
 \begin{subfigure}[h]{0.95\textwidth}
     \centering
     \includegraphics[width=\textwidth]{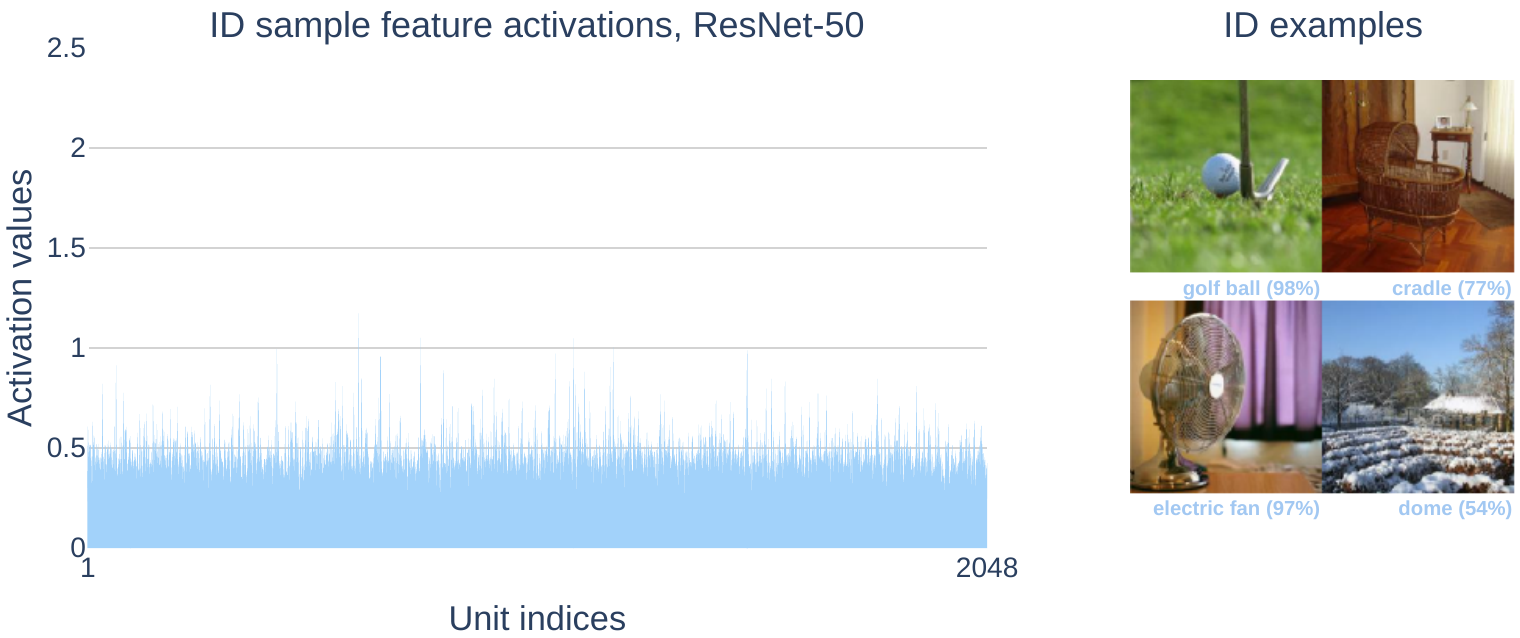}
 \end{subfigure}
 \begin{subfigure}[h]{0.95\textwidth}
     \centering
     \includegraphics[width=\textwidth]{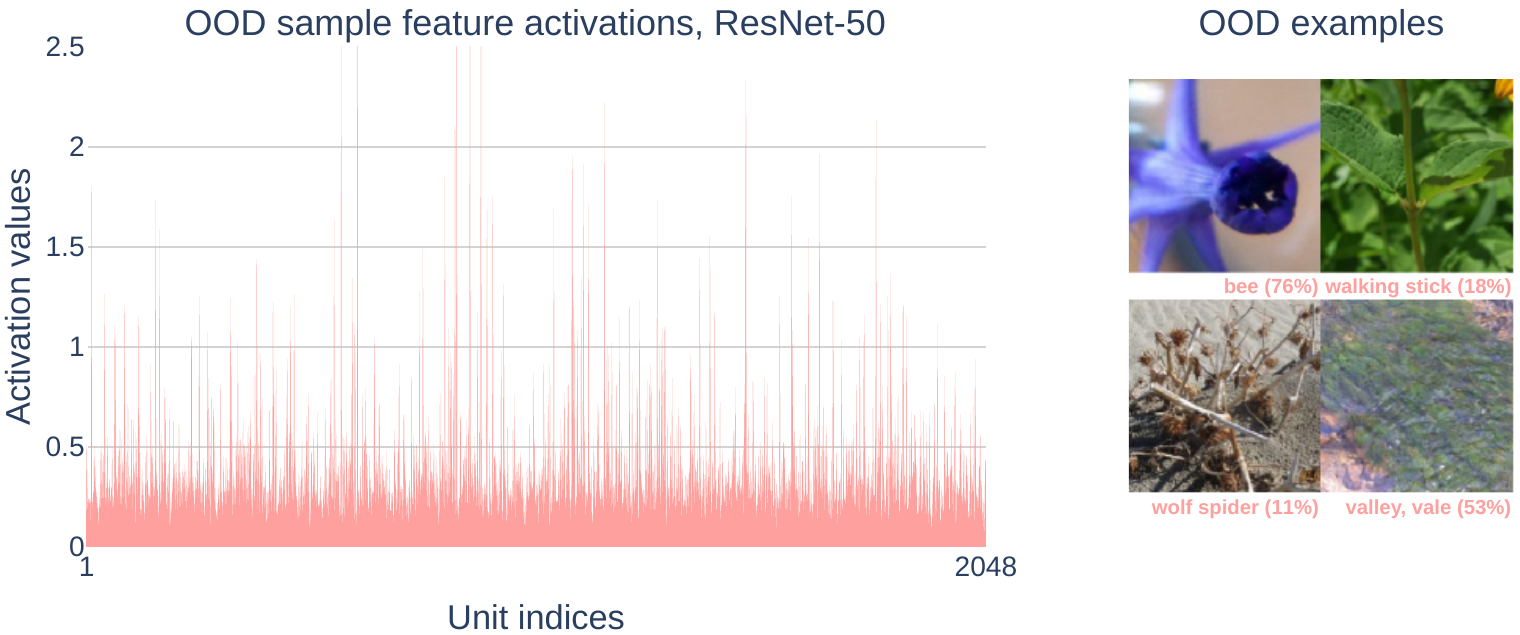}
 \end{subfigure}
\caption{\textbf{Activation values and examples of ID and OOD samples.} On the left we plot the activation values of all the 2048 units in the penultimate layer of a ResNet-50 pretrained on ImageNet-1k, of ID (ImageNet-1k) and OOD (iNaturalist) samples. 100 samples are taken from each dataset and the values are their average. The figure is a replication of Figure 1(b) of~\citet{react}. On the right we show example pictures from the corresponding dataset, including class prediction and confidence.}
\figlabel{intuition}
\end{figure*}

\section{Experiments}

In this section, we describe the experimental setup used to evaluate our method. We first introduce the benchmarks, covering a diverse range of in-distribution and out-of-distribution datasets across different scales and domains. We then outline the evaluation metrics and finally, we present experimental results that demonstrate the strong performance and generalizability of LTS, with particular attention to its applicability across variety of architectures.

\subsection{OOD detection benchmark}

We tested our method on four different benchmarks presented in \tabref{benchmarks}.
\begin{table*}[h]
\centering
\resizebox{0.99\textwidth}{!}{
\centering
\begin{tabular}{ l | l }
\toprule

\multirow{3}{*}{CIFAR-10 benchmark} & \textbf{ID Dataset:} CIFAR-10 \\
& \textbf{OOD Datasets:} SVHN, LSUN C, LSUN R, iSUN, Places365, Textures \\
& \textbf{Model architecture:} DenseNet-101\\
\midrule

\multirow{3}{*}{CIFAR-100 benchmark} & \textbf{ID Dataset:} CIFAR-100 \\
& \textbf{OOD Datasets:} SVHN, LSUN C, LSUN R, iSUN, Places365, Textures \\
& \textbf{Model architecture:} DenseNet-101\\

\midrule

\multirow{3}{*}{OpenOOD benchmark} & \textbf{ID Dataset:} ImageNet-1k \\
& \textbf{OOD Datasets:} SSB-hard, NINCO (Near OOD), iNaturalist, Textures, OpenImage-O (Far OOD) \\
& \textbf{Model architecture:} ResNet50\\
\midrule
\multirow{3}{*}{ImageNet benchmark} & \textbf{ID Dataset:} ImageNet-1k \\
& \textbf{OOD Datasets:} iNaturalist, SUN, Places365, Textures \\
& \textbf{Model architectures:} ResNet50, MobileNetV2, ViT-B-16, ViT-L-16, SWIN-S, SWIN-B, MLP-B, MLP-L\\
\bottomrule
\end{tabular}
}
\caption{\textbf{Benchmarks used in our OOD experiments.} We evaluate our method on 4 different OOD benchmarks covering small and large scale datasets. Our tests include evaluations on 3 ID dataset, 14 OOD datasets and 9 architectures.
}
\tablabel{benchmarks}
\end{table*}

\textbf{CIFAR-10 and CIFAR-100 benchmarks.} The setup for CIFAR-10 and CIFAR-100 benchmarks is derived from \citet{djurisic2022extremely, react, sun2022dice} and includes 6 OOD datasets: SVHN~\citep{netzer2011reading}, LSUN-Crop~\citep{yu2015lsun}, LSUN-Resize~\citep{yu2015lsun}, iSUN~\citep{xu2015turkergaze}, Places365~\citep{zhou2017places} and Textures~\citep{cimpoi2014describing}. We used DenseNet-101~\citep{huang2017densely} pre-trained on corresponding ID dataset.

\textbf{OpenOOD benchmark.} This benchmark, introduced by ~\citet{zhang2023openood}, is designed to rigorously evaluate the effectiveness of out-of-distribution detection algorithms by providing a diverse set of datasets. OpenOOD suite divides the OOD datasets into two categories: Near-OOD and Far-OOD. The Near-OOD datasets comprise SSB-hard~\citep{vaze2021open} and NINCO~\citep{bitterwolf2023or} whereas Far-OOD dataset include iNaturalist~\citep{van2018inaturalist}, Textures~\citep{cimpoi2014describing} and OpenImage-O~\citep{wang2022vim}. Experiments for these tasks were conducted in alignment with setup from ~\citet{xu2023scaling} and we used ResNet50~\citep{he2016deep} architecture pre-trained on ImageNet-1k~\citep{krizhevsky2017imagenet}.

\textbf{ImageNet benchmark.} The tests utilize setup derived from~\citet{djurisic2022extremely, xu2023scaling, zhao2024towards}. ID dataset is ImageNet-1k and OOD datasets include iNaturalist~\citep{van2018inaturalist}, SUN~\citep{xiao2010sun}, Places365~\citep{zhou2017places} and Textures~\citep{cimpoi2014describing}. Recent work by \citet{djurisic2022extremely} and \citet{xu2023scaling}, evaluated ImageNet benchmark on 2 architectures ResNet50~\citep{he2016deep}, MobileNetV2~\citep{sandler2018mobilenetv2} while more recent study by~\citet{zhao2024towards} extends the study to 8 different architectures. We adopted the same set of architectures and performed ImageNet benchmark evaluation on: ResNet50~\citep{he2016deep}, MobileNetV2~\citep{sandler2018mobilenetv2}, ViT-B-16, ViT-L-16~\citep{dosovitskiy2020image}, SWIN-S, SWIN-B~\citep{liu2021swin}, MLP-B and MLP-L~\citep{tolstikhin2021mlp}.

\subsection{OOD evaluation metrics}
We evaluate our method using standard OOD detection metrics following the work of~\citet{hendrycks17baseline}: AUROC (Area Under the Receiver Operating Characteristic Curve), FPR@95 (False Positive Rate at $95\%$ True Positive Rate) and AUPR (Area Under the Precision-Recall curve). AUROC measures model's ability to differentiate between ID and OOD classes. Higher AUROC values indicate better model performance. FPR@95 measures the proportion of false positives (incorrectly identified as positive) out of the total actual negatives, at the threshold where the true positive rate (correctly identified positives) is $95\%$. Lower FPR@95 indicates better model performance. AUPR is providing a single scalar value that reflects the model's ability to balance precision and recall. A higher AUPR indicates a better performing model.

\subsection{OOD detection performance}
LTS demonstrates excellent performance in out-of-distribution detection. As highlighted in~\tabref{cifar_results}, the results on the CIFAR-10 and CIFAR-100 benchmarks show that LTS outperforms the previously leading method in both evaluation metrics. Detail CIFAR-10 and CIFAR-100 results are provided in Appendix \ref{appendix_cifar}.
\begin{table}[h]
\centering 
\resizebox{\linewidth}{!}{
\begin{tabular}{l l l | l l}
\hline
 & \multicolumn{2}{c}{\textbf{CIFAR-10}} & \multicolumn{2}{c}{\textbf{CIFAR-100}} \\
\textbf{Method} & \textbf{FPR95} & \textbf{AUROC} & \textbf{FPR95} & \textbf{AUROC} \\
 & {\quad $\downarrow$} & \quad $\uparrow$ & \quad $\downarrow$ & \quad $\uparrow$ \\
\hline \\
Softmax score & 48.73 & 92.46 & 80.13 & 74.36 \\
ODIN & 24.57 & 93.71 & 58.14 & 84.49 \\
Mahalanobis & 31.42 & 89.15 & 55.37 & 82.73 \\
Energy score & 26.55 & 94.57 & 68.45 & 81.19 \\
ReAct & 26.45 & 94.95 & 62.27 & 84.47 \\
DICE & ${20.83}^{\pm1.58}$ & ${95.24}^{\pm0.24}$ & ${49.72}^{\pm1.69}$ & ${87.23}^{\pm0.73}$ \\
ASH-P & 23.45 & 95.22 & 64.53 & 82.71\\
ASH-B & 20.23 & 96.02 & 48.73 & 88.04\\
ASH-S & 15.05 & 96.61 & 41.40 & 90.02\\
SCALE & 12.57 & 97.27 & 38.99 & 90.74 \\
\rowcolor{lightgray}\textbf{LTS (Ours)} & \textbf{12.13} & \textbf{97.40} & \textbf{37.55} & \textbf{90.78}\\
\bottomrule
\end{tabular}
}
\caption{\textbf{OOD detection results on CIFAR benchmarks.} LTS enhances the state-of-the-art performance across all evaluation metrics on the CIFAR benchmarks. The results are averaged across 6 OOD tasks. $\uparrow$ indicates that higher values are better, while a $\downarrow$ signifies that lower values are preferable. All values are presented as percentages. Table results, except for LTS (indicated as "Ours") and SCALE, are sourced from~\citet{djurisic2022extremely}.}
\tablabel{cifar_results}
\todo{optfs}
\andrija{We copyed everything from ASH, and re-run new experiments with the same setup for LTS and SCALE}
\end{table}

\begin{table*}[h]
\resizebox{\linewidth}{!}{
\begin{tabular}{c c c c c c c | c c c c c c c c c}
\hline
\\
 & \multicolumn{6}{c}{\textbf{Near OOD}} & \multicolumn{8}{c}{\textbf{Far OOD}} \\
 \\
\textbf{Methods} & \multicolumn{2}{c}{\textbf{SSB-hard}} & \multicolumn{2}{c}{\textbf{Ninco}} & \multicolumn{2}{c}{\textbf{Average}}  & \multicolumn{2}{c}{\textbf{iNaturalist}} & \multicolumn{2}{c}{\textbf{Textures}} & \multicolumn{2}{c}{\textbf{OpenImage-O}} & \multicolumn{2}{c}{\textbf{Average}} \\
& & & & & & & & & & & & & & & \\
& FPR95 & AUROC & FPR95 & AUROC & FPR95 & AUROC  & FPR95 & AUROC & FPR95 & AUROC & FPR95 & AUROC & FPR95 & AUROC \\
& $\downarrow$ & $\uparrow$ & $\downarrow$ & $\uparrow$ & $\downarrow$ & $\uparrow$  & $\downarrow$ & $\uparrow$ & $\downarrow$ & $\uparrow$ & $\downarrow$ & $\uparrow$ & $\downarrow$ & $\uparrow$ \\
\midrule

\multicolumn{1}{l}{Softmax score} & 74.49 & 72.09 & 56.84 & 79.95 & 65.67 & 76.02  & 43.34 & 88.41 & 60.89 & 82.43 & 50.16 & 84.86 & 51.47 & 85.23 \\
\multicolumn{1}{l}{Mahalanobis} & 76.19 & 72.51 & 59.49 & 80.41 & 67.84 & 76.46  & 30.63 & 91.16 & 46.11 & 88.39 & 37.86 & 89.17 & 38.20 & 89.58 \\
\multicolumn{1}{l}{Energy score} & 76.54 & 72.08 & 60.59 & 79.70 & 68.56 & 75.89  & 31.33 & 90.63 & 45.77 & 88.7 & 38.08 & 89.06 & 38.40 & 89.47 \\
\multicolumn{1}{l}{ReAct} & 77.57 & 73.02 & 55.92 & 81.73 & 66.75 & 77.38  & 16.73 & 96.34 & 29.63 & 92.79 & 32.58 & 91.87 & 26.31 & 93.67 \\
\multicolumn{1}{l}{ASH-S} & 70.80 & 74.72 & 53.26 & 84.54 & 62.03 & 79.63  & 11.02 & 97.72 & 10.90 & 97.87 & 28.60 & 93.82 & 16.86 & 96.47 \\
\multicolumn{1}{l}{SCALE} & 67.72 & 77.35 & 51.80 & \textbf{85.37} & 59.76 & \textbf{81.36}  & 9.51 & 98.02 & \textbf{11.90} & \textbf{97.63} & \textbf{28.18} & \textbf{93.95} & \textbf{16.53} & \textbf{96.53} \\
\rowcolor{lightgray} \multicolumn{1}{l}{\textbf{LTS (Ours)} } & \textbf{67.36} & \textbf{77.55} & \textbf{51.15} & 85.16 & \textbf{59.26} & 81.35  & \textbf{9.34} & \textbf{98.06} & 12.10 & 97.58 & 29.21 & 93.77  & 16.88 & 96.47 \\

\bottomrule
\end{tabular}
}
\caption{\textbf{LTS performance on OpenOOD benchmark.} The model used for these results is ResNet-50 pretrained on ImageNet-1k (ID) dataset. On the NearOOD task, LTS performs comparably to the previous state-of-the-art and slightly improves it in terms of the FPR@95 evaluation metric. On the FarOOD task, LTS is the second best model in terms of both metrics. $\uparrow$ indicates larger values are better and $\downarrow$ indicates smaller values are better. All values are percentages. Method results except for LTS (marked as ``Ours'') are taken from~\citet{xu2023scaling}.}
\tablabel{openood_results}
\todo{optfs}
\andrija{We copied everything from SCALE, and re-run new experiments with the same setup for LTS}
\end{table*}

\begin{table*}[h]
\resizebox{\linewidth}{!}{
\begin{tabular}{c c c c c c c c c c c c}
\hline
 & & \multicolumn{8}{c}{\textbf{OOD Datasets}} & & \\
\textbf{Model} & \textbf{Methods} & \multicolumn{2}{c}{\textbf{iNaturalist}} & \multicolumn{2}{c}{\textbf{SUN}} & \multicolumn{2}{c}{\textbf{Places}} & \multicolumn{2}{c}{\textbf{Textures}} & \multicolumn{2}{c}{\textbf{Average}} \\
& & & & & & & & & & & \\
& & FPR95 & AUROC & FPR95 & AUROC & FPR95 & AUROC & FPR95 & AUROC & FPR95 & AUROC \\
& & $\downarrow$ & $\uparrow$ & $\downarrow$ & $\uparrow$ & $\downarrow$ & $\uparrow$ & $\downarrow$ & $\uparrow$ & $\downarrow$ & $\uparrow$ \\
\midrule

\multirow{12}{*}{ResNet50} & \multicolumn{1}{l}{Softmax score} & 54.99 & 87.74 & 70.83 & 80.86 & 73.99 & 79.76 & 68.00 & 79.61 & 66.95 & 81.99 \\
& \multicolumn{1}{l}{ODIN} & 47.66 & 89.66 & 60.15 & 84.59 & 67.89 & 81.78 & 50.23 & 85.62 & 56.48 & 85.41 \\
& \multicolumn{1}{l}{Mahalanobis} & 97.00 & 52.65 & 98.50 & 42.41 & 98.40 & 41.79 & 55.80 & 85.01 & 87.43 & 55.47 \\
& \multicolumn{1}{l}{Energy score} & 55.72 & 89.95 & 59.26 & 85.89 & 64.92 & 82.86 & 53.72 & 85.99 & 58.41 & 86.17 \\
& \multicolumn{1}{l}{ReAct} & 20.38 & 96.22 & 24.20 & 94.20 & 33.85 & 91.58 & 47.30 & 89.80 & 31.43 & 92.95 \\
& \multicolumn{1}{l}{DICE} & 25.63 & 94.49 & 35.15 & 90.83 & 46.49 &  87.48 & 31.72 & 90.30 & 34.75 & 90.77 \\
& \multicolumn{1}{l}{ASH-P} & 44.57 & 92.51 & 52.88 & 88.35 & 61.79 & 85.58 & 42.06 & 89.70 & 50.32 & 89.04 \\
 & \multicolumn{1}{l}{ASH-B} & 14.21 & 97.32 & 22.08 & 95.10 & 33.45 & 92.31 & 21.17 & 95.50 & 22.73 & 95.06 \\
 & \multicolumn{1}{l}{ASH-S } & 11.49 & 97.87 & 27.98 & 94.02 & 39.78 & 90.98 & \textbf{11.93} & \textbf{97.60} & 22.80 & 95.12 \\
 & \multicolumn{1}{l}{ASH-B+ATS} & 24.07 & 95.19 & 32.70 & 92.37 & 45.63 &  88.33 & 18.71 & 96.29 & 30.28 & 93.05 \\
& \multicolumn{1}{l}{OptFS (V)} & 18.33 & 96.63 & 37.03 & 92.84 & 45.97 & 90.15 & 24.80 & 95.48 & 31.53 & 93.78 \\
& \multicolumn{1}{l}{OptFS (S)} & 15.90 & 97.00 & 34.00 & 93.28 & 43.61 & 90.50 & 21.61 & 95.99 & 28.78 &  94.19 \\
& \multicolumn{1}{l}{SCALE} & 9.50 & 98.17 & 23.27 & 95.02 & 34.51 & 92.26 & 12.93  & 97.37 & 20.05 & 95.71 \\
\rowcolor{lightgray} & \multicolumn{1}{l}{\textbf{LTS (Ours)} } & \textbf{9.44} & \textbf{98.17} & \textbf{22.04} & \textbf{95.30} & \textbf{32.92} & \textbf{92.79} & 15.27 & 96.81 & \textbf{19.92} & \textbf{95.77} \\

\midrule
\multirow{12}{*}{MobileNet} & \multicolumn{1}{l}{Softmax score} & 64.29 & 85.32 & 77.02 & 77.10 & 79.23 & 76.27 & 73.51 & 77.30 & 73.51 & 79.00 \\
& \multicolumn{1}{l}{ODIN} & 55.39 & 87.62 & 54.07 & 85.88 & 57.36 & 84.71 & 49.96 & 85.03 & 54.20 & 85.81 \\
& \multicolumn{1}{l}{Mahalanobis} & 62.11 & 81.00 & 47.82 & 86.33 & 52.09 & 83.63 & 92.38 & 33.06 & 63.60 & 71.01 \\
& \multicolumn{1}{l}{Energy score} & 59.50 & 88.91 & 62.65 & 84.50 & 69.37 & 81.19 & 58.05 & 85.03 & 62.39 & 84.91 \\
& \multicolumn{1}{l}{ReAct} & 42.40 & 91.53 & 47.69 & 88.16 & 51.56 & 86.64 & 38.42 & 91.53 & 45.02 & 89.47 \\
& \multicolumn{1}{l}{DICE} & 43.09 &  90.83 & 38.69 & 90.46 & 53.11 & 85.81 & 32.80 & 91.30 & 41.92 & 89.60 \\

 & \multicolumn{1}{l}{ASH-P} & 54.92 & 90.46 & 58.61 & 86.72 & 66.59 & 83.47 & 48.48 & 88.72 & 57.15 & 87.34 \\
 & \multicolumn{1}{l}{ASH-B} & 31.46 & 94.28 & 38.45 & 91.61 & 51.80 & 87.56 & 20.92 & 95.07 & 35.66 & 92.13\\
 & \multicolumn{1}{l}{ASH-S} & 39.10 & 91.94 & 43.62 & 90.02 & 58.84 & 84.73 & 13.12 & 97.10 & 38.67 & 90.95 \\
& \multicolumn{1}{l}{SCALE} & 38.20 & 91.67 & 42.64 & 90.09 & 58.73 & 84.00 & \textbf{12.59}  & \textbf{97.47} & 38.04 & 90.81 \\ 
\rowcolor{lightgray} & \multicolumn{1}{l}{\textbf{LTS (Ours)}} & \textbf{29.78} & \textbf{94.33} & \textbf{36.14} & \textbf{92.34} & \textbf{50.65} & \textbf{87.84} & 14.02 & 96.89 & \textbf{32.65} & \textbf{92.85} \\ 
\bottomrule
\end{tabular}
}
\caption{\textbf{OOD detection results on ImageNet-1k benchmark.} LTS outperforms all existing methods on this benchmark. $\uparrow$ indicates that higher values are better, while $\downarrow$ signifies that lower values are preferable. Results indicated "Ours" are computed by us. ATS results are copied from \citet{krumpl2024ats}. ATS performs best when combined with ASH-B, so we have included their best results for reference. The remaining values in the table are sourced from Table 1 in~\citet{djurisic2022extremely}.}
\andrija{ASH-B + ATS is copied from ATS paper, OptFS (V) and OptFS(S) are taken from OptFS paper, SCALE and LTS are performed by us with the same setup. The rest is copied from ASH.}
\tablabel{imagenet_results}
\end{table*}

On the OpenOOD benchmark, LTS achieves performance comparable to the previous state-of-the-art method, SCALE~\citep{xu2023scaling}. More specifically, on Near-OOD benchmark LTS marginally improves FPR@95 while on Far-OOD benchmark its performance is slightly below that of the previous state-of-the-art. \tabref{openood_results} showcase LTS results on OpenOOD benchmark.

\tabref{imagenet_results} showcases the performance of LTS on the ImageNet-1k benchmark across two architectures, ResNet-50 and MobileNetV2. LTS surpasses the previous benchmarks in both metrics across all OOD tasks, with the exception of ImageNet-1k (ID) vs Textures task (OOD). On average, LTS sets a new state-of-the-art, offering improvements in both metrics for ResNet-50. For MobileNetV2, it achieves a reduction in FPR@95 by $3\%$ while enhancing the AUROC evaluation metric.

\subsection{LTS applicability across architectures}

A recent study by \citet{zhao2024towards} revealed that many previous methods fail to maintain robust out-of-distribution performance across different architectures. Specifically, modern transformer-based architectures such as Vision Transformer~\citep{dosovitskiy2020image}, Swin Transformer~\citep{liu2021swin}, as well as MLP-Mixer~\citep{tolstikhin2021mlp} present significant challenges for existing OOD detection methods. \mladen{Unlike other architectures ReLU, Vit, Swin, MLP Mixer} The OptFS~\citep{zhao2024towards} method was the first to demonstrate consistent performance across eight different architectures. In our experiments, we replicate the exact experimental setup of OptFS and demonstrate that our method performs consistently well across all tested architectures, significantly reducing FPR@95 while maintaining a comparable level of AUROC compared to OptFS. \figref{fig:arch_comp} illustrates the performance of LTS across five different architectures. Detailed results for all tested architectures can be found in Appendix  \ref{appendix_arch}.
Integration of LTS within the different architectures is described in Appendix \ref{appendix_vit}.

\begin{figure*}[hbtp]
 \centering
 \begin{subfigure}[h]{\textwidth}
    \centering 
	\includegraphics[width=\textwidth]{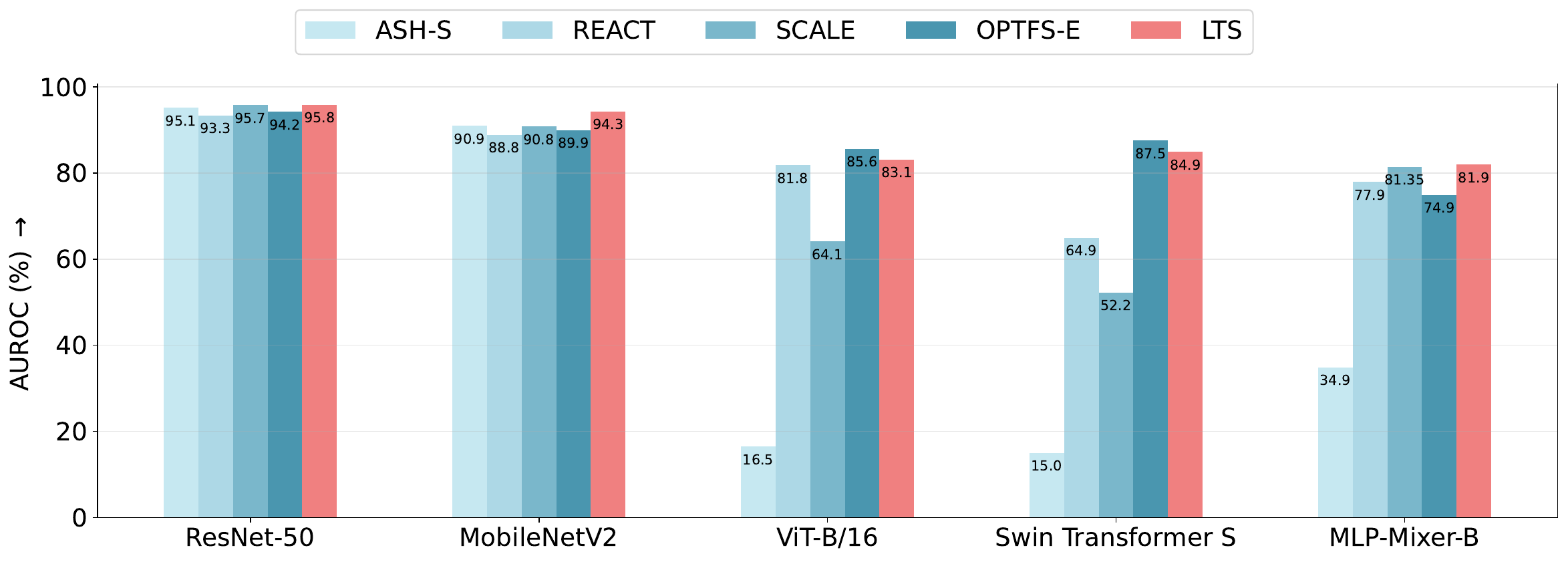}
 \end{subfigure}
 \hfill
 \begin{subfigure}[h]{\textwidth}
    \centering 
	\includegraphics[width=\textwidth]{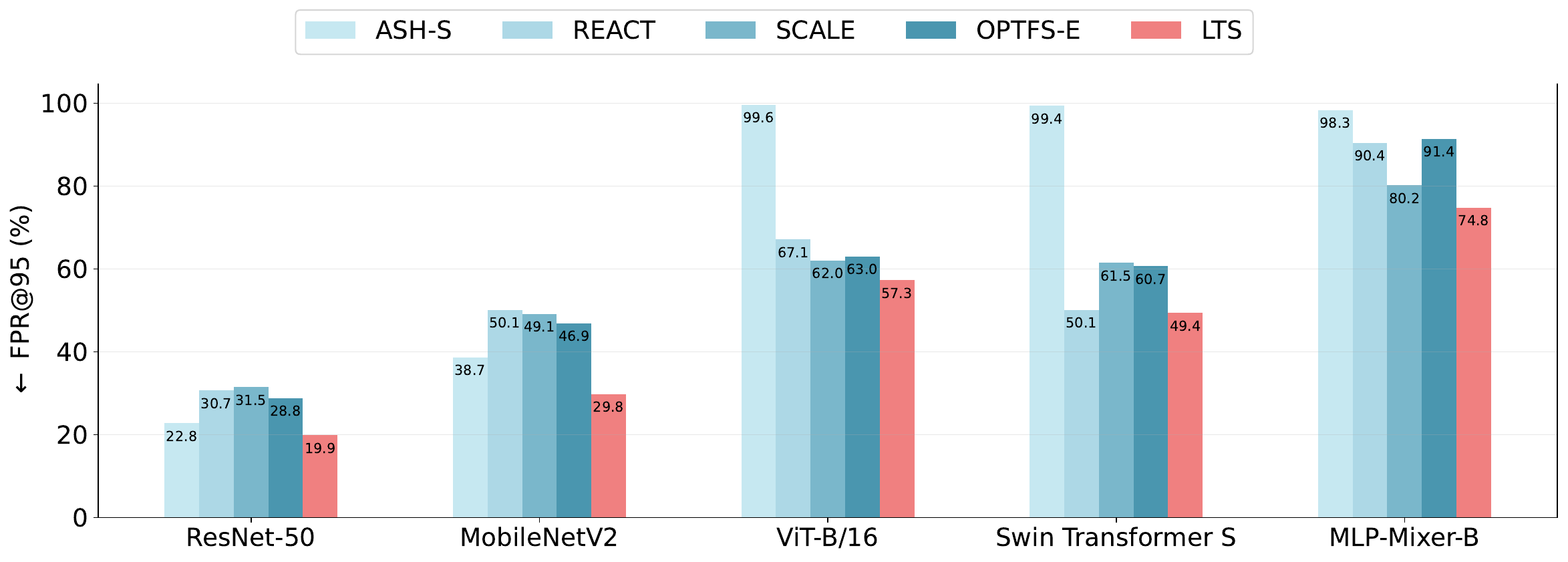}
 \end{subfigure}
\caption{\textbf{Performance comparison of OOD methods across five architectures.} This figure compares the performance of various OOD detection methods across five different architectures. The evaluation is based on two metrics: AUROC (top figure) and FPR@95 (bottom figure). All results are tested on ImageNet-1k benchmark and averaged across 4 tasks (iNaturalist, SUN, Places, and Textures). Higher AUROC indicates better performance, while lower FPR@95 indicates better performance.} 
\figlabel{fig:arch_comp}
\end{figure*}

\subsection{Failure cases and limitations}

LTS achieves state-of-the-art results overall but slightly underperforms SCALE on the relatively easy Far-OOD benchmark, where both methods achieve comparable performance. We hypothesize that SCALE’s advantage arises from its aggressive exponential scaling, which can be particularly effective on OOD samples that are substantially different from the in-distribution data. Appendix~\ref{appendix_lts_failure_cases} presents visualizations of LTS failure cases, illustrating challenges such as subtle domain shifts, texture bias, and high semantic overlap between ID and OOD samples.

Non-convolutional architectures such as Vision Transformers, Swin Transformers, and MLP-Mixers present additional challenges for OOD detection. In these models, the penultimate layer features tend to be more complex and less structured compared to convolutional networks, due to their global receptive fields and attention-based mechanisms. This added complexity makes distinguishing between in-distribution and out-of-distribution samples more difficult. As shown in ~\figref{fig:arch_comp}, we observe slight degradation in performance, specifically an increase in the false positive rate, when applying LTS to non-convolutional architectures. We present LTS failure cases on ViT-B and Swin-B architectures in Appendix~\ref{appendix_lts_failure_cases}. Addressing these challenges and improving OOD detection on such architectures will be an important direction for our future work.

Apart from these empirically observed challenges, there are other potential limitations of our method and other similar methods. Namely, the energy score relies on meaningful information conveyed via logits. In the ideal case, ID samples will have a large positive logit, say $f_i(\mathbf{x})$ for one class, and large negative logits for other classes. In extreme, the energy score is close to $-f_i(\mathbf{x})$. On the other hand, OOD samples will have all negative logits (since no class is the appropriate one). In extreme, the energy score is a large positive number. However, these properties depend on the quality of the analyzed model, specifically the quality of its penultimate layer activations. For instance, if the model confuses an OOD sample for a representative of some class and assigns it a high logit, the energy score might be low or negative and thus deceiving. If the model cannot separate some classes, their probabilities tend to be conservative (in order to avoid large errors which the cross-entropy penalizes severely), resulting in low logits, which might lead to higher values of the energy score even for ID samples. In conclusion, various practical obstacles during training, such as noisy data or extreme class imbalance, can reduce the quality of the analyzed model. These might also negatively impact the OOD detection performance of energy-score-based methods (including ours) for that model.

\subsection{Ablation studies}

\textbf{Determining the optimal value of $p$.} We examine how varying the hyperparameter $p$, which specifies the proportion of top activations used to compute the scaling factor $S$, affects the performance of our method.

The optimal value of $p$ depends on both, the underlying architecture and ID dataset. To assess the effect of $p$, we conduct a comprehensive evaluation across five architectures: ResNet-50, MobileNet-V2, ViT-B/16, Swin Transformer-S, and MLP-Mixer-B, using four tasks from the ImageNet-1k benchmark. Additionally, we perform a complementary experiment in which the architecture is fixed to DenseNet, and performance is evaluated across three different in-distribution datasets: CIFAR-10, CIFAR-100, and ImageNet-1k. Across all experiments, our empirical results consistently indicate that setting $p = 5\%$ yields strong performance on all metrics. A detailed illustration is presented in \figref{choose_p}.

\begin{figure*}[h]
 \centering
 Estimating optimal threshold $p$ across different architectures\\[0.3em]
 \begin{subfigure}[h]{0.49\textwidth}
     \centering
     \includegraphics[width=\textwidth]{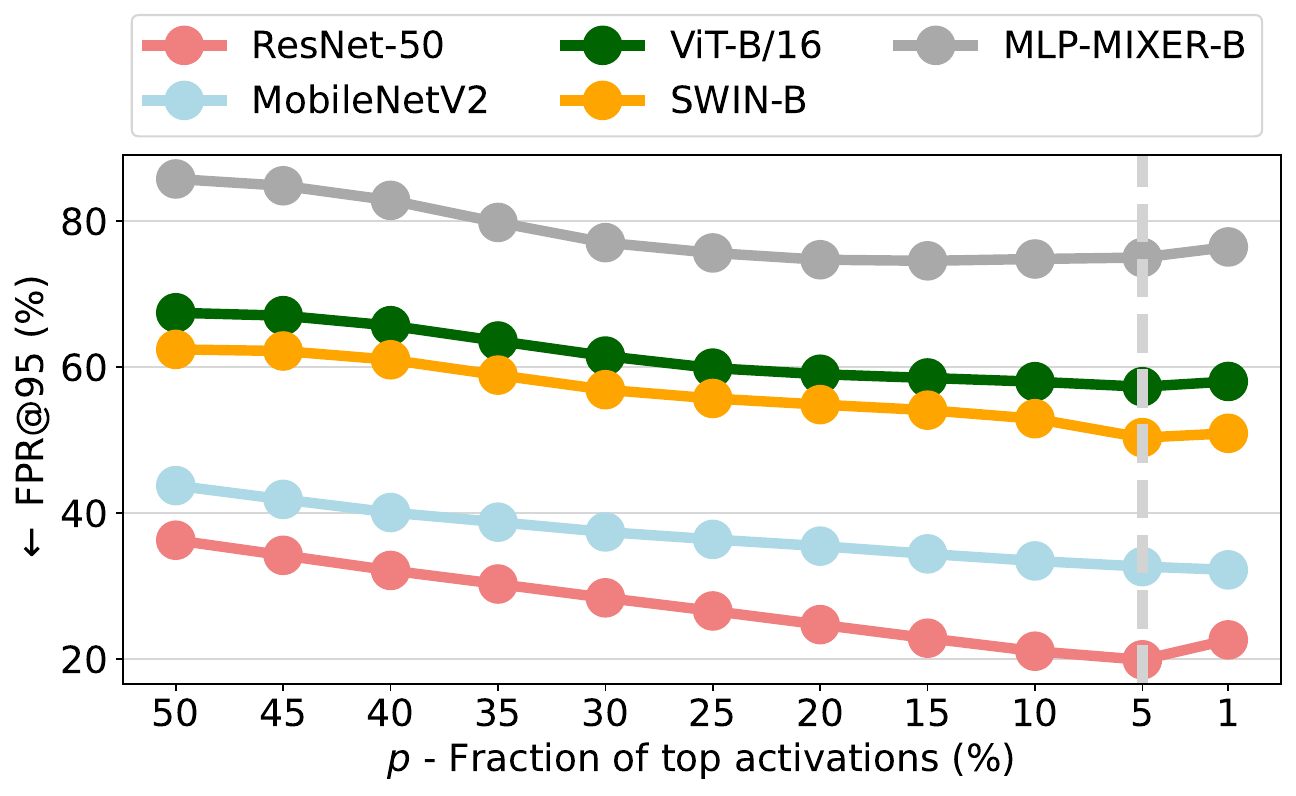}
 \end{subfigure}
 \begin{subfigure}[h]{0.49\textwidth}
     \centering
     \includegraphics[width=\textwidth]{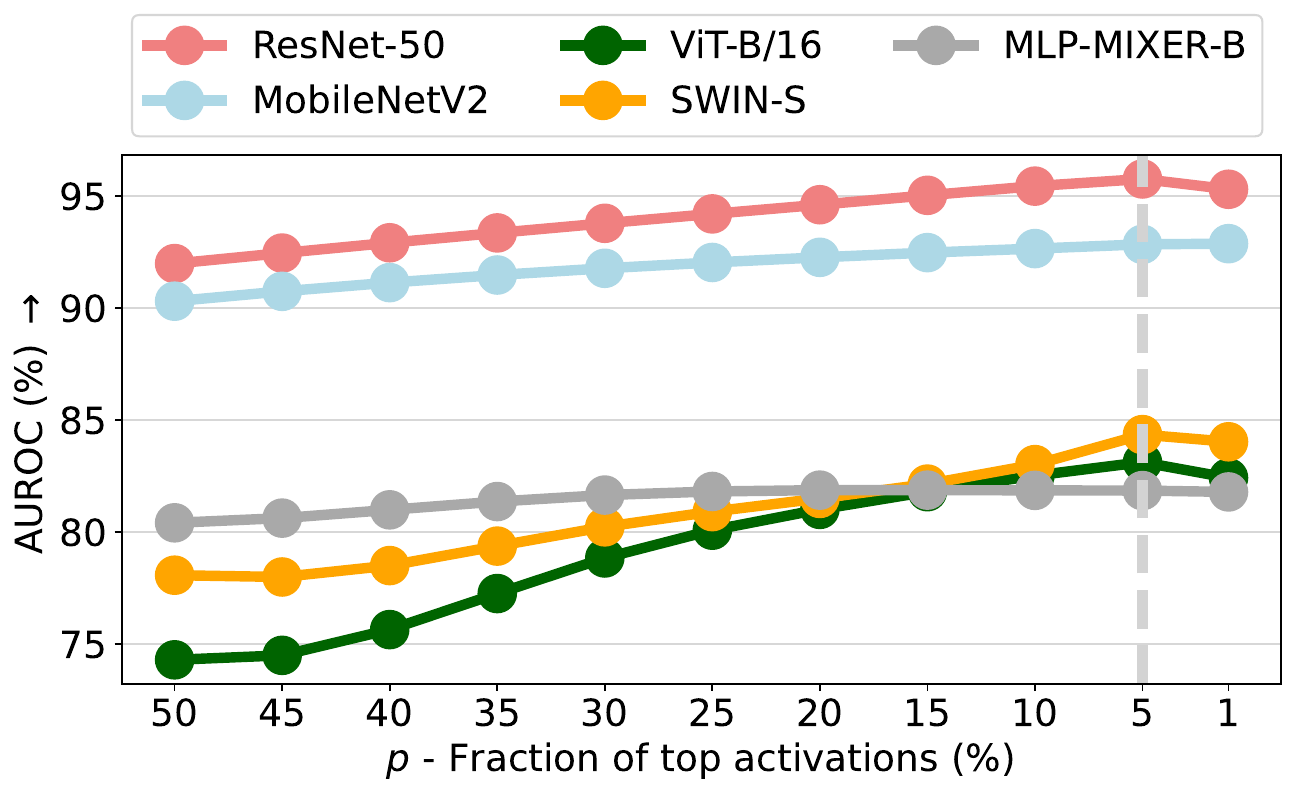}
 \end{subfigure}
\vspace{0.5cm}

  Estimating optimal threshold $p$ across different ID datasets on DenseNet architecture\\[0.3em]
 \begin{subfigure}[h]{0.49\textwidth}
     \centering
     \includegraphics[width=\textwidth]{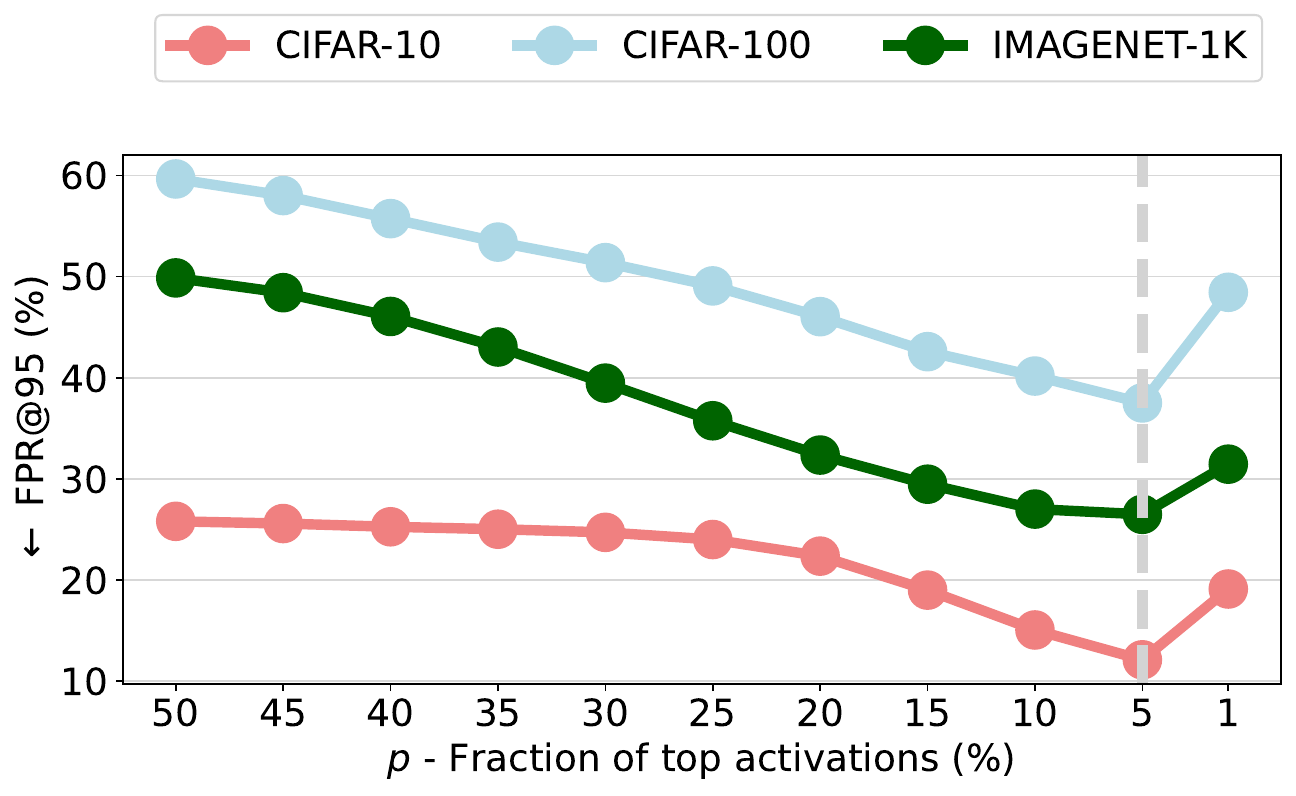}
     
 \end{subfigure}
 \begin{subfigure}[h]{0.49\textwidth}
     \centering
     \includegraphics[width=\textwidth]{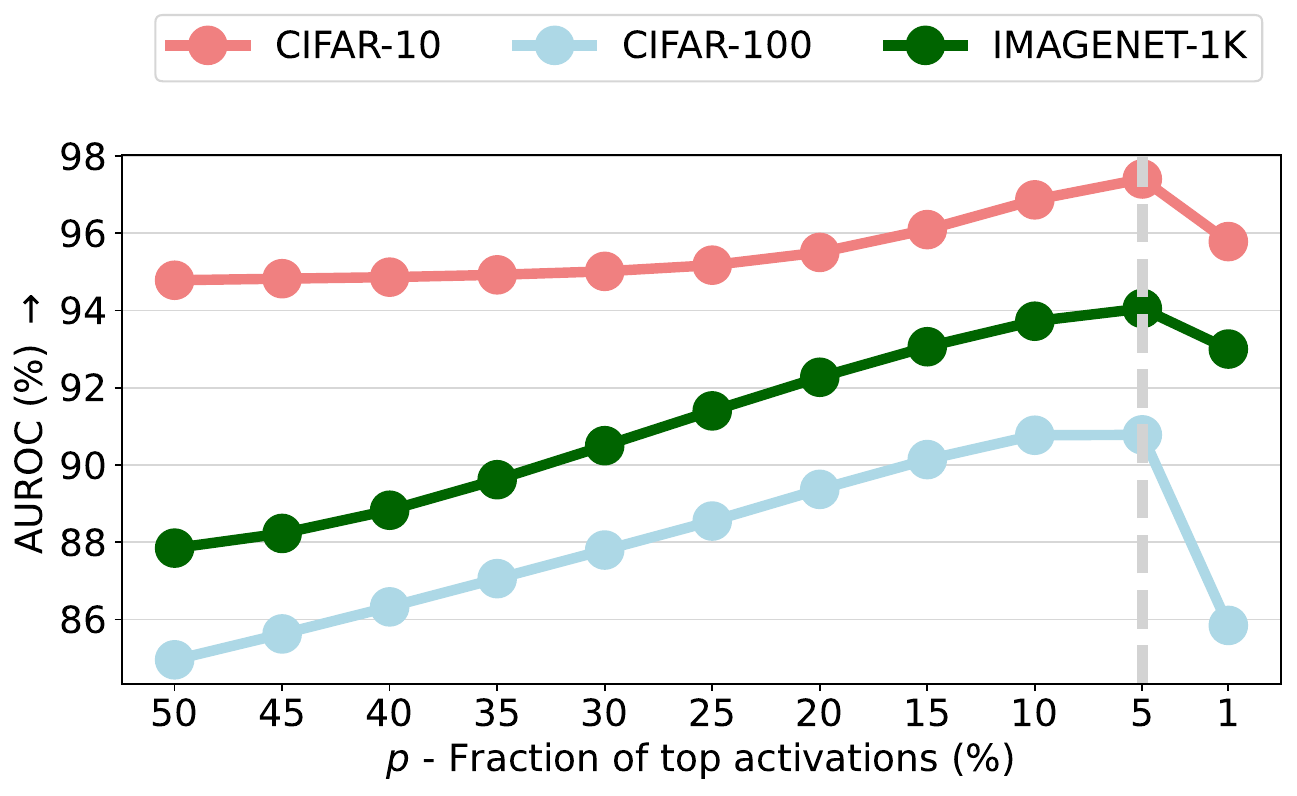}
 \end{subfigure}

\caption{\textbf{Analysis of optimal $p$ values for scaling factor calculation.} The top row shows the performance of LTS evaluated across five architectures using various values of the hyperparameter $p$, with results averaged over four tasks. The bottom row focuses on a single architecture, namely DenseNet, pretrained on three different in-distribution datasets: CIFAR-10, CIFAR-100, and ImageNet-1k. Across all our experiments, LTS consistently achieves optimal performance at $p = 5\%$ on both evaluation metrics, AUROC and FPR@95. We therefore recommend using $p = 5\%$ as the default setting.}
\figlabel{choose_p}
\end{figure*}

\textbf{Compatibility with other OOD detection methods.} 
LTS is not compatible with methods that scale or prune activations in intermediate layers, such as ASH-S~\citep{djurisic2022extremely} and SCALE~\citep{xu2023scaling}, due to its operational characteristics.

In contrast, ReAct~\citep{react} complements LTS effectively. ReAct clips penultimate layer activations to distinguish in-distribution from out-of-distribution data. In our approach, we first compute the scaling factor $S$ using LTS, apply ReAct’s clipping, and finally scale the logits using $S$. \tabref{react_lts} shows that this combination consistently improves ReAct’s performance across all evaluation metrics.

\begin{table}[hbtp]
\centering
\resizebox{\linewidth}{!}{
\begin{tabular}{c c c c c}
\hline
 & & & & \\
\textbf{Model} & \textbf{Methods} & \multicolumn{3}{c}{\textbf{Average}} \\
& & & & \\
& & FPR95 $\downarrow$ & AUROC $\uparrow$ & AUPR $\uparrow$ \\
\midrule
\multirow{2}{*}{ResNet-50} & \multicolumn{1}{l}{ReAct} & 30.72 & 93.27 & 98.59\\
 & \multicolumn{1}{l}{\textbf{ReAct + LTS}} &  \textbf{19.74} & \textbf{95.77} & \textbf{99.08}\\
\midrule
\multirow{2}{*}{MobileNet} & \multicolumn{1}{l}{ReAct} & 48.95 & 88.75 & 97.46\\
 & \multicolumn{1}{l}{\textbf{ReAct + LTS}} & \textbf{33.57} & \textbf{92.59} & \textbf{98.29} \\
\midrule
\multirow{2}{*}{ViT-B/16} & \multicolumn{1}{l}{ReAct} & 64.99 & 80.74 & 94.65\\
 & \multicolumn{1}{l}{\textbf{ReAct + LTS}} & \textbf{57.59} & \textbf{84.24} & \textbf{96.05} \\
\midrule
\multirow{2}{*}{Swin Transformer S} & \multicolumn{1}{l}{ReAct} & 67.47 & 72.26 & 90.93\\
& \multicolumn{1}{l}{\textbf{ReAct + LTS}} & \textbf{61.97} & \textbf{79.99} & \textbf{94.58} \\
\bottomrule
\end{tabular}
}

\caption{\textbf{Compatibility of LTS and ReAct.} ReAct significantly benefits when combined with LTS. Our approach involves first using LTS to calculate the scaling factor $S$, then applying the ReAct rectification operation, and finally scaling the logits using $S$. Presented results are averaged across 4 tasks on ImageNet benchmark.}
\tablabel{react_lts}
\end{table}

\section{Conslusion}

In this study, we introduced LTS, extremely simple, post-hoc, off-the-shelf method for detecting out-of-distribution samples. LTS operates by deriving a scaling factor for each sample based on activations from the penultimate layer, which is then applied to adjust the logits. We conducted thorough testing of LTS, demonstrating that its performance surpasses many existing methods. Additionally, we have shown its robustness and effectiveness across a diverse set of architectures. Looking ahead, we plan to explore strategies to consistently maintain performance on specific OOD tasks irrespective of architectural differences.

\clearpage

\appendix



\section{Detailed CIFAR-10 And CIFAR-100 Results}\label{appendix_cifar}
\tabref{detailresultscifar10} and \tabref{detailresultscifar100} supplement \tabref{cifar_results} in the main text, as they display the full results on each of the 6 OOD datasets for DenseNet-101 trained on CIFAR-10 and CIFAR-100 respectively.

\begin{sidewaystable*}
\centering
\resizebox{\textwidth}{!}{
\begin{tabular}{lllllllllllllll} \toprule
\multirow{2}{*}{\textbf{Method}} & \multicolumn{2}{c}{\textbf{SVHN}} & \multicolumn{2}{c}{\textbf{LSUN-c}} & \multicolumn{2}{c}{\textbf{LSUN-r}} & \multicolumn{2}{c}{\textbf{iSUN}} & \multicolumn{2}{c}{\textbf{Textures}} & \multicolumn{2}{c}{\textbf{Places365}} & \multicolumn{2}{c}{\textbf{Average}} \\ 
 & \textbf{FPR95} & \textbf{AUROC} & \textbf{FPR95} & \textbf{AUROC} & \textbf{FPR95} & \textbf{AUROC} & \textbf{FPR95} & \textbf{AUROC} & \textbf{FPR95} & \textbf{AUROC} & \textbf{FPR95} & \textbf{AUROC} & \textbf{FPR95} & \textbf{AUROC} \\ 
 & \quad $\downarrow$ & \quad $\uparrow$ & \quad $\downarrow$ & \quad $\uparrow$ & \quad  $\downarrow$ & \quad $\uparrow$ & \quad $\downarrow$ & \quad  $\uparrow$ &  \quad $\downarrow$ & \quad $\uparrow$ & \quad $\downarrow$ & \quad $\uparrow$ & \quad $\downarrow$ & \quad $\uparrow$ \\ \midrule
Softmax score  & 47.24 & 93.48 & 33.57 & 95.54 & 42.10 & 94.51 & 42.31 & 94.52 & 64.15 & 88.15 & 63.02 & 88.57 & 48.73 & 92.46 \\
ODIN  & 25.29 & 94.57 & 4.70 & 98.86 & 3.09 & 99.02 & 3.98 & 98.90 & 57.50 & 82.38 & 52.85 & 88.55 & 24.57 & 93.71 \\
GODIN  & 6.68 & 98.32 & 17.58 & 95.09 & 36.56 & 92.09 & 36.44 & 91.75 & 35.18 & 89.24 & 73.06 & 77.18 & 34.25 & 90.61 \\
Mahalanobis  & 6.42 & 98.31 & 56.55 & 86.96 & 9.14 & 97.09 & 9.78 & 97.25 & 21.51 & 92.15 & 85.14 & 63.15 & 31.42 & 89.15 \\
Energy score & 40.61 & 93.99 & 3.81 & 99.15 & 9.28 & 98.12 & 10.07 & 98.07 & 56.12 & 86.43 & 39.40 & 91.64 & 26.55 & 94.57 \\ 
ReAct  & 41.64 & 93.87 & 5.96 & 98.84 & 11.46 & 97.87 & 12.72 & 97.72 & 43.58 & 92.47 & 43.31 & 91.03 & 26.45 & 94.67 \\
 {DICE} 
 & 25.99$^{\pm{5.10}}$ & 95.90$^{\pm{1.08}}$ & 0.26$^{\pm{0.11}}$ & 99.92$^{\pm{0.02}}$ & 3.91$^{\pm{0.56}}$ & 99.20$^{\pm{0.15}}$ & 4.36$^{\pm{0.71}}$ & 99.14$^{\pm{0.15}}$ & 41.90$^{\pm{4.41}}$ & 88.18$^{\pm{1.80}}$ & 48.59$^{\pm{1.53}}$ & 89.13$^{\pm{0.31}}$ & 20.83$^{\pm{1.58}}$ & 95.24$^{\pm{0.24}}$ \\
 
 ASH-P & 30.14 & 95.29 & 2.82 & 99.34 & 7.97 & 98.33 & 8.46 & 98.29 & 50.85 & 88.29 & 40.46 & 91.76 & 23.45 & 95.22\\
ASH-B & 17.92 & 96.86 & 2.52 & 99.48 & 8.13 & 98.54 & 8.59 & 98.45 & 35.73 & 92.88 & 48.47 & 89.93 & 20.23 & 96.02 \\
ASH-S & 6.51 & 98.65 & 0.90 & 99.73 & 4.96 & 98.92 & 5.17 & 98.90 & 24.34 & 95.09 & 48.45 & 88.34 & 15.05 & 96.61 \\
SCALE & 5.77 & 98.72 & \textbf{0.72} & \textbf{99.74} &  \textbf{3.35} & \textbf{99.22} & \textbf{3.44} & \textbf{99.21} & 23.39 & 94.97 & 38.61 & 91.74 & 12.55 & 97.27 \\
\rowcolor{lightgray}\textbf{LTS (Ours)} & \textbf{5.54} & \textbf{98.81} & 0.74 & \textbf{99.74} & 3.49 & 99.21 & 3.70 & 99.20 & \textbf{21.44} & \textbf{95.56} & \textbf{37.84} & \textbf{91.89} & \textbf{12.12} & \textbf{97.40} \\
\bottomrule
\end{tabular}
}
\caption{\small Detailed results on six common OOD benchmark datasets: Textures~\citep{cimpoi2014describing}, SVHN~\citep{netzer2011reading}, Places365~\citep{zhou2017places}, LSUN-Crop~\citep{yu2015lsun}, LSUN-Resize~\citep{yu2015lsun}, and iSUN~\citep{xu2015turkergaze}. For each ID dataset, we use the same DenseNet-101 pretrained on \textbf{CIFAR-10}. $\uparrow$ indicates larger values are better and $\downarrow$ indicates smaller values are better.}
\tablabel{detailresultscifar10}
\end{sidewaystable*}

\begin{sidewaystable*} 
\centering
\resizebox{\textwidth}{!}{
\begin{tabular}{lllllllllllllll} \toprule
\multirow{2}{*}{\textbf{Method}} & \multicolumn{2}{c}{\textbf{SVHN}} & \multicolumn{2}{c}{\textbf{LSUN-c}} & \multicolumn{2}{c}{\textbf{LSUN-r}} & \multicolumn{2}{c}{\textbf{iSUN}} & \multicolumn{2}{c}{\textbf{Textures}} & \multicolumn{2}{c}{\textbf{Places365}} & \multicolumn{2}{c}{\textbf{Average}} \\ 

 & \textbf{FPR95} & \textbf{AUROC} & \textbf{FPR95} & \textbf{AUROC} & \textbf{FPR95} & \textbf{AUROC} & \textbf{FPR95} & \textbf{AUROC} & \textbf{FPR95} & \textbf{AUROC} & \textbf{FPR95} & \textbf{AUROC} & \textbf{FPR95} & \textbf{AUROC} \\ 
 & \quad $\downarrow$ & \quad $\uparrow$ & \quad $\downarrow$ & \quad $\uparrow$ & \quad  $\downarrow$ & \quad $\uparrow$ & \quad $\downarrow$ & \quad  $\uparrow$ &  \quad $\downarrow$ & \quad $\uparrow$ & \quad $\downarrow$ & \quad $\uparrow$ & \quad $\downarrow$ & \quad $\uparrow$ \\ \midrule
Softmax score & 81.70 & 75.40 & 60.49 & 85.60 & 85.24 & 69.18 & 85.99 & 70.17 & 84.79 & 71.48 & 82.55 & 74.31 & 80.13 & 74.36 \\
ODIN & 41.35 & 92.65 & 10.54 & 97.93 & 65.22 & 84.22 & 67.05 & 83.84 & 82.34 & 71.48 & 82.32 & 76.84 & 58.14 & 84.49 \\
GODIN & 36.74 & 93.51 & 43.15 & 89.55 & 40.31 & 92.61 & 37.41 & 93.05 & 64.26 & 76.72 & 95.33 & 65.97 & 52.87 & 85.24 \\ 
Mahalanobis & 22.44 & 95.67 & 68.90 & 86.30 & \textbf{23.07} & \textbf{94.20} & \textbf{31.38} & \textbf{93.21} & 62.39 & 79.39 & 92.66 & 61.39 & 55.37 & 82.73 \\
Energy score & 87.46 & 81.85 & 14.72 & 97.43 & 70.65 & 80.14 & 74.54 & 78.95 & 84.15 & 71.03 & 79.20 & 77.72 & 68.45 & 81.19 \\ 
ReAct & 83.81 & 81.41 & 25.55 & 94.92 & 60.08 & 87.88 & 65.27 & 86.55 & 77.78 & 78.95 & 82.65 & 74.04 & 62.27 & 84.47 \\
{DICE} & 54.65$^{\pm{4.94}}$ & 88.84$^{\pm{0.39}}$ & 0.93$^{\pm{0.07}}$ & 99.74$^{\pm{0.01}}$ & 49.40$^{\pm{1.99}}$ & 91.04$^{\pm{1.49}}$ & 48.72$^{\pm{1.55}}$ & 90.08$^{\pm{1.36}}$ & 65.04$^{\pm{0.66}}$ & 76.42$^{\pm{0.35}}$ & 79.58$^{\pm{2.34}}$ & 77.26$^{\pm{1.08}}$ & 49.72$^{\pm{1.69}}$ & 87.23$^{\pm{0.73}}$ \\

ASH-P & 81.86 & 83.86 & 11.60 & 97.89 & 67.56 & 81.67 & 70.90 & 80.81 & 78.24 & 74.09 & \textbf{77.03} & \textbf{77.94} & 64.53 & 82.71\\
ASH-B & 53.52 & 90.27 & \textbf{4.46} & \textbf{99.17} & 48.38 & 91.03 & 47.82 & 91.09 & 53.71 & 84.25 & 84.52 & 72.46 & 48.73 & 88.04 \\
ASH-S & 25.02 & 95.76 & 5.52 & 98.94 & 51.33 & 90.12 & 46.67 & 91.30 & 34.02 & 92.35 & 85.86 & 71.62 & 41.40 & 90.02 \\
SCALE & 24.35 & 94.99 & 18.24 & 96.09 & 53.61 & 87.71 & 46.66 & 89.43 & \textbf{30.25} & 91.16 & 93.64 & 54.65 & 44.46 & 85.67 \\
\rowcolor{lightgray}\textbf{LTS (Ours)} & \textbf{21.57} & \textbf{96.11} &  4.89 & 99.07 & 43.29 & 91.92 & 38.35 & 93.02 & 31.12 & \textbf{93.15} & 85.84 & 71.41 & \textbf{37.51} & \textbf{90.78}\\
\bottomrule
\end{tabular}
}
\caption{\small Detailed results on six common OOD benchmark datasets: Textures~\citep{cimpoi2014describing}, SVHN~\citep{netzer2011reading}, Places365~\citep{zhou2017places}, LSUN-Crop~\citep{yu2015lsun}, LSUN-Resize~\citep{yu2015lsun}, and iSUN~\citep{xu2015turkergaze}. For each ID dataset, we use the same DenseNet-101 pretrained on \textbf{CIFAR-100}. $\uparrow$ indicates larger values are better and $\downarrow$ indicates smaller values are better. }

\tablabel{detailresultscifar100}
\end{sidewaystable*}

\section{LTS Performance Across Eight Architectures}\label{appendix_arch}
\tabref{architectures} supplements \figref{fig:arch_comp} and presents detailed performance of LTS across 8 different architectures on ImageNet benchmark, along with other methods.

\begin{sidewaystable*}
\centering
\resizebox{\textwidth}{!}{
\begin{tabular}{lcccccccccccccccccc}
\toprule
\multirow{2}{*}{\textbf{Method}} & \multicolumn{2}{c}{\textbf{ResNet50}} & \multicolumn{2}{c}{\textbf{MobileNetV2}} & \multicolumn{2}{c}{\textbf{ViT-B-16}} & \multicolumn{2}{c}{\textbf{ViT-L-16}} & \multicolumn{2}{c}{\textbf{SWIN-S}} & \multicolumn{2}{c}{\textbf{SWIN-B}} & \multicolumn{2}{c}{\textbf{MLP-B}} & \multicolumn{2}{c}{\textbf{MLP-L}} & \multicolumn{2}{c}{\textbf{Average}}\\
& \textbf{FPR@95}$\downarrow$ & \textbf{AUROC}$\uparrow$ & \textbf{FPR@95}$\downarrow$ & \textbf{AUROC}$\uparrow$ & \textbf{FPR@95}$\downarrow$ & \textbf{AUROC}$\uparrow$ & \textbf{FPR@95}$\downarrow$ & \textbf{AUROC}$\uparrow$ & \textbf{FPR@95}$\downarrow$ & \textbf{AUROC}$\uparrow$ & \textbf{FPR@95}$\downarrow$ & \textbf{AUROC}$\uparrow$ & \textbf{FPR@95}$\downarrow$ & \textbf{AUROC}$\uparrow$ & \textbf{FPR@95}$\downarrow$ & \textbf{AUROC}$\uparrow$  & \textbf{FPR@95}$\downarrow$ & \textbf{AUROC}$\uparrow$\\
\midrule
MSP & 64.76 & 82.82 & 70.47 & 80.67 & 61.74 & 83.12 & 65.22 & 81.75 & 59.68 & 83.75 & 62.79 & 81.38 & 69.36 & 81.97 & 76.01 & 80.04 & 66.25 & 81.94\\
ODIN & 57.68 & 87.08 & 60.42 & 86.39 & 61.24 & 79.59 & \textbf{64.06} & 78.65 & 57.30 & 80.99 & 64.36 & 73.77 & 63.05 & 85.15 & 73.17 & 82.29 & 62.66 & 81.74\\
Energy & 57.47 & 87.05 & 58.87 & 86.59 & 67.41 & 74.31 & 68.43 & 74.65 & 62.82 & 77.65 & 75.32 & 64.87 & 85.99 & 79.96 & 84.44 & 79.38 & 70.08 & 78.06\\
DICE & 35.72 & 90.92 & 41.93 & 89.60 & 90.30 & 70.77 & 71.77 & 67.12 & 88.68 & 39.70 & \textbf{40.71} & 59.71 & 83.36 & 62.63 & 81.53 & 75.40 & 67.55 & 70.49\\
ReAct & 30.70 & 93.30 & 50.09 & 88.81 & 67.08 & 81.78 & 59.68 & 83.50 & 50.09 & 87.02 & 64.86 & 83.24 & 90.41 & 77.86 & 80.71 & 79.13 & 63.73 & 84.44\\
BFAct &  31.41 & 92.98 & 48.35 & 89.19 & 73.26 & 82.75 & 81.16 & 82.69 & 57.20 & \textbf{88.21} & 65.44 & \textbf{86.62} & 96.76 & 67.46 & 96.36 & 72.16 & 68.74 & 82.76 \\
ASH-P &  50.33 & 89.04 & 57.15 & 87.34 & 99.45 & 20.30 & 99.42 & 18.37 & 99.11 & 19.21 & 99.08 & 20.59 & 98.77 & 36.88 & 99.04 & 29.20 & 87.79 & 40.11 \\
ASH-B &  22.73 & 95.06 & 35.66 & 92.13 & 94.33 & 49.14 & 94.08 & 37.89 & 96.42 & 30.00 & 91.47 & 47.38 & 99.83 & 19.14 & 66.05 & 84.31 & 75.07 & 56.88 \\
ASH-S &  22.80 & 95.12 & 38.67 & 90.95 & 99.62 & 16.47 & 99.55 & 16.72 & 99.36 & 14.98 & 99.35 & 17.75 & 98.31 & 34.85 & 99.36 & 18.65 & 82.13 & 38.18 \\
VRA-P &  25.49 & 94.57 & 45.53 & 89.85 & 98.02 & 34.64 & 99.62 & 14.99 & 99.38 & 18.15 & 99.65 & 15.51 & 99.64 & 16.58 & 99.23 & 19.49 & 83.32 & 37.97 \\
SCALE & 20.05 & 95.71 & 38.04 & 90.81 & 78.34 & 64.08 & 92.77 & 49.25 & 91.95 & 52.21 &  79.07 & 68.75 & 80.62 & 81.35 & 96.87 & 68.91 & 72.21 & 71.38\\
OptFS(V)&  31.53 & 93.78 & 49.09 & 89.62 & 62.01 & \textbf{85.80} & 68.61 & \textbf{85.10} & 61.47 & 87.04 & 62.96 & 86.09 & 80.23 & 78.83 & 82.59 & 78.92 & 62.31 & 85.65 \\
OptFS(E)&  28.78 & 94.19 & 46.92 & 89.90 & 63.04 & 85.64 & 74.96 & 84.69 & 60.67 & 87.50 & 63.50 & 86.55 & 91.37 & 74.85 & 88.49 & 78.22 & 64.71 & 85.19 \\
 \rowcolor{lightgray}\textbf{LTS (Ours)} & \textbf{19.92} & \textbf{95.77} & \textbf{29.78} & \textbf{94.33} & \textbf{57.31} & 83.12 & 65.05 & 75.33 & \textbf{49.43} & 84.87 & 63.97 & 78.57 & 74.80 & 81.86 & 78.01 & 79.69 & \textbf{54.78} & 84.19\\
\\
\bottomrule
\end{tabular}
}%
\caption{OOD detection results on ImageNet benchmark across 8 model architectures. OOD datasets are iNaturalist, SUN, Places, and Textures. $\uparrow$ indicates larger values are better and $\downarrow$ indicates smaller values are better.}
\tablabel{architectures}
\end{sidewaystable*}

\section{Application of LTS to different architectures}\label{appendix_vit}
\figref{vit_resnet_pos} illustrates the integration of LTS within the Vision Transformer and ResNet-50 architectures.
We observe that ResNet-50, DenseNet-101 and MobileNetV2 have non-negative activations at penultimate layer, unlike ViT, Swin Transformer and MLP-Mixer. To mimic that, in OOD detection time we apply ReLU on penultimate layer activations of ViT, Swin Transformer and MLP-Mixer before feeding them into LTS. We emipirically observed that such a modification leads to increase in OOD detection performance.

\begin{figure*}[h]
\centering
\includegraphics[width=\textwidth]{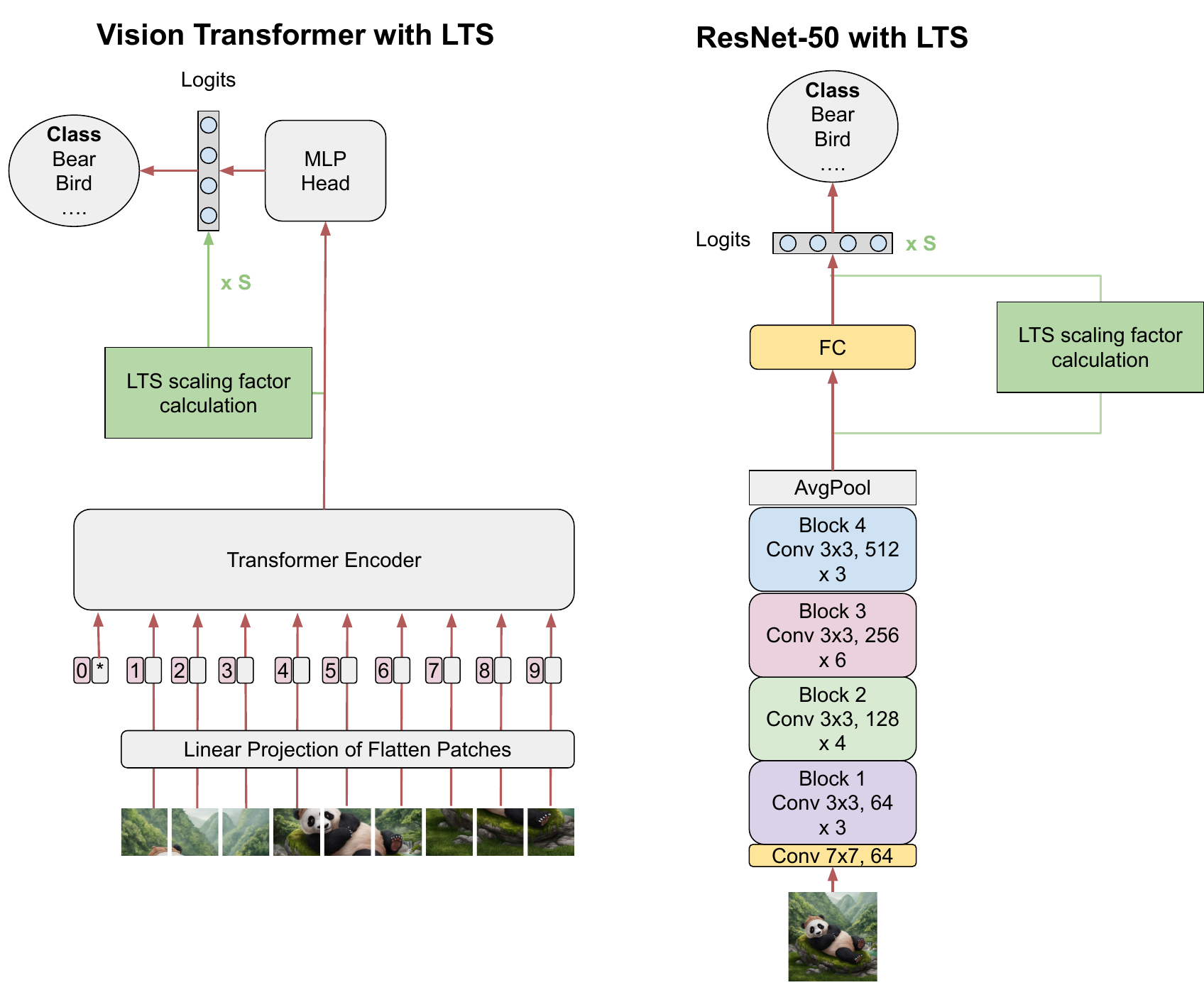}
\caption{\textbf{Integration of LTS in Vision Transformer (ViT) and ResNet-50 architectures.} The diagrams illustrate the placement of the LTS module, which calculates scaling factor used to adjust the logits before applying OOD detection scoring function. In ViT, LTS is applied before the MLP Head, following the Transformer Encoder's output. In ResNet-50, LTS is added post average pooling layer and before the fully connected layer.}
\figlabel{vit_resnet_pos}
\end{figure*}

\section{Differences between LTS and SCALE}
\label{appendix_diff_lts_vs_scale}
SCALE modifies the penultimate layer activations by applying a scaling factor during the forward pass, which consequently alters the network’s internal representations and can impact network's performance on original task. In contrast, LTS computes a scaling factor during the forward pass but applies it only to the logits, preserving the original activations and maintaining network's performance. Notably, the scaling factor in SCALE follows an exponential form, $e^\frac{S1}{S2}$, a strategy first introduced by ASH-S, whereas LTS adopts a simpler quadratic scaling, using $(\frac{S1}{S2})^2$. Empirical results demonstrate that LTS outperforms SCALE across the majority of architectures and benchmarks.

\section{Differences between LTS and ATS}
\label{appendix_diff_lts_vs_ats}
LTS and ATS both aim to improve OOD detection by scaling logits, but differ fundamentally in their methods. ATS relies on training data to build an empirical CDF and computes a sample-specific temperature for the energy score, whereas LTS requires no training data and imposes no distributional assumptions. Unlike ATS, which adjusts the energy score via temperature scaling, LTS directly scales the logits before energy computation. Empirically, ATS performs best when combined with other methods, while LTS is designed to be stand-alone and achieves significantly stronger results independently.

\section{Hyperparameter $p$ and ID vs. OOD separation}
\label{appendix_lts_intuition}

An effective way to illustrate how LTS separates ID and OOD data is to observe the changes in energy score distributions as the hyperparameter $p$ is varied. Since the goal is to shrink the overlap between the two, we track the intersection over union (IoU) as a natural metric. In~\figref{morphing}, we follow this process: starting from no treatment (first plot) and moving toward a cleaner separation between ID and OOD as $p$ decreases.

\begin{figure*}[h]
\centering
\includegraphics[width=.3\textwidth]{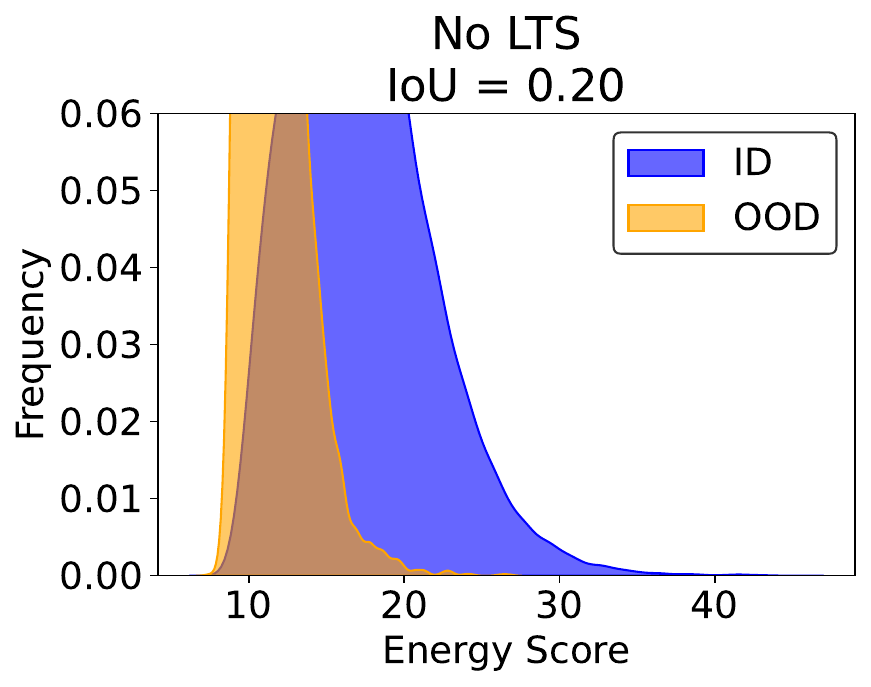}
\includegraphics[width=.3\textwidth]{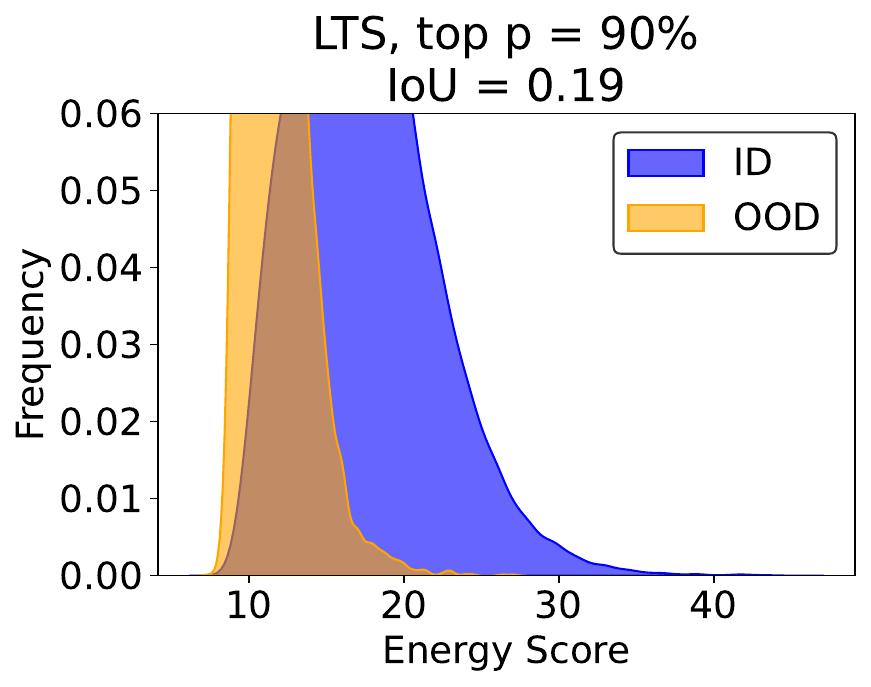}
\includegraphics[width=.3\textwidth]{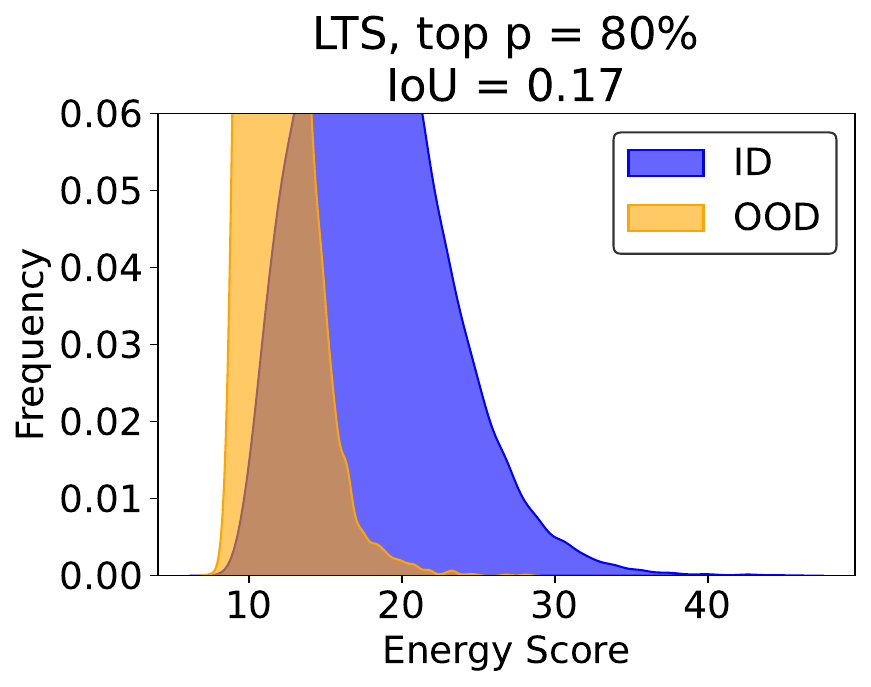}
\includegraphics[width=.3\textwidth]{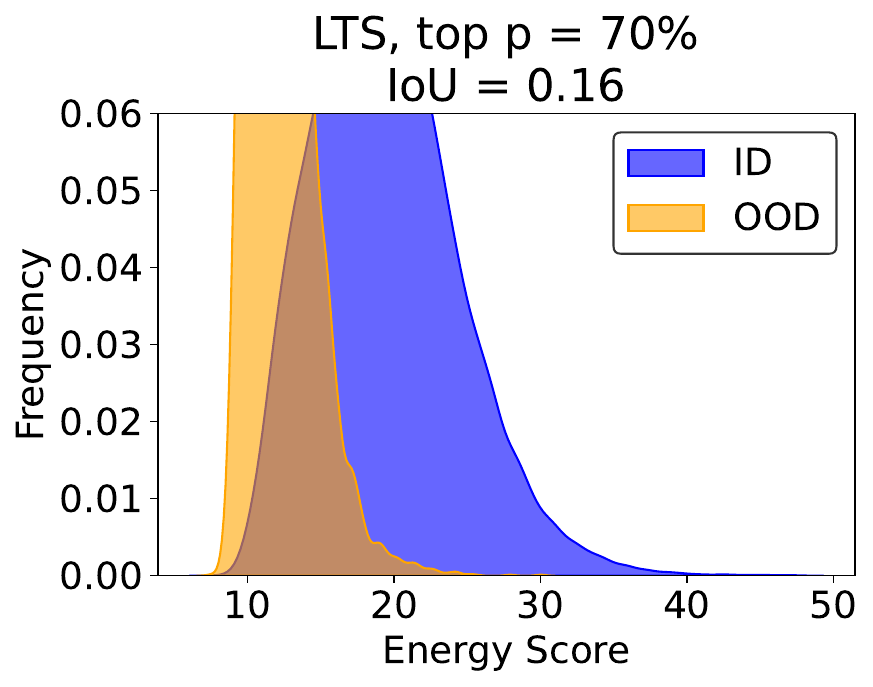}
\includegraphics[width=.3\textwidth]{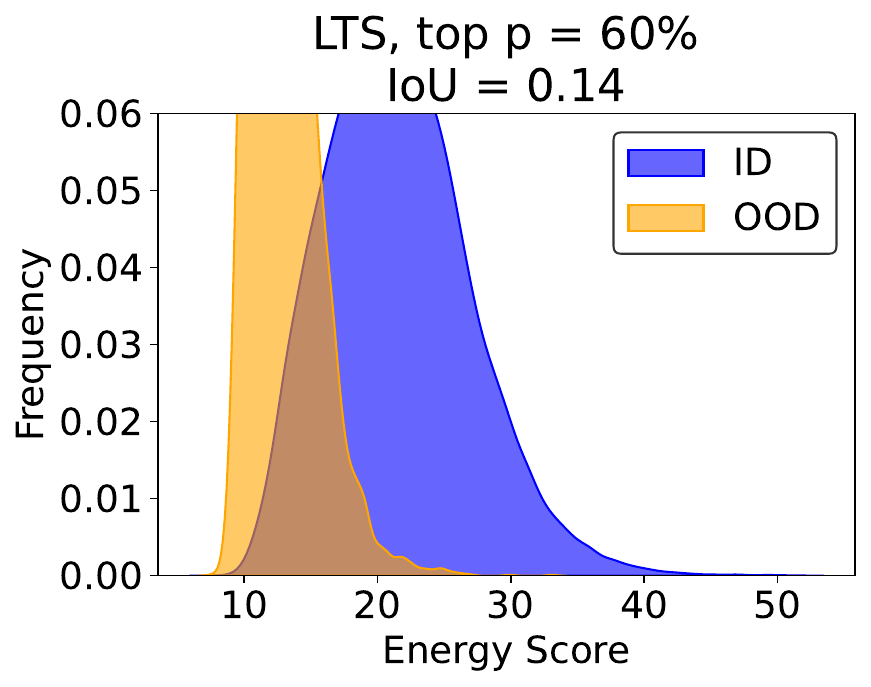}
\includegraphics[width=.3\textwidth]{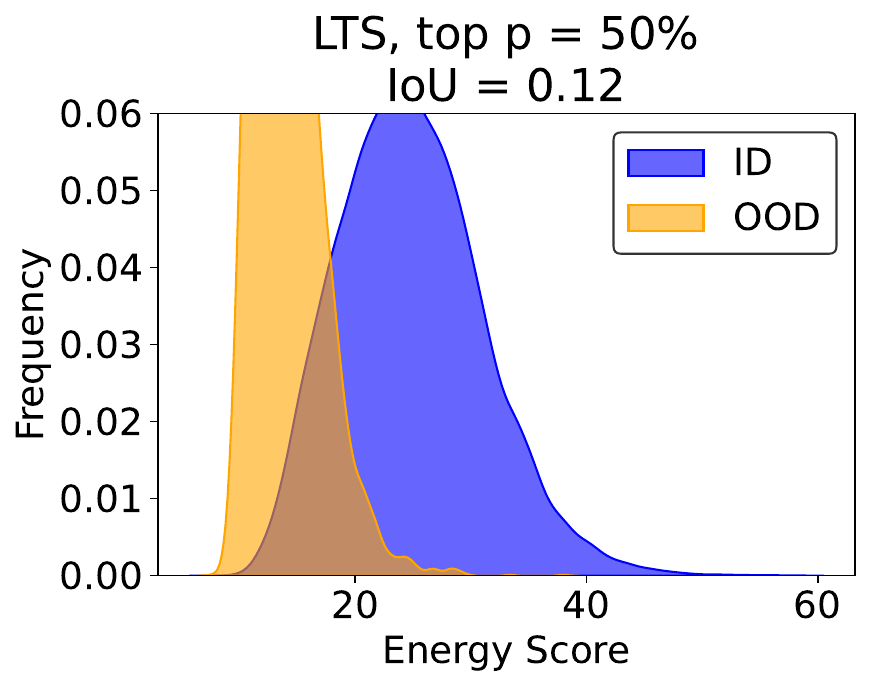}
\includegraphics[width=.3\textwidth]{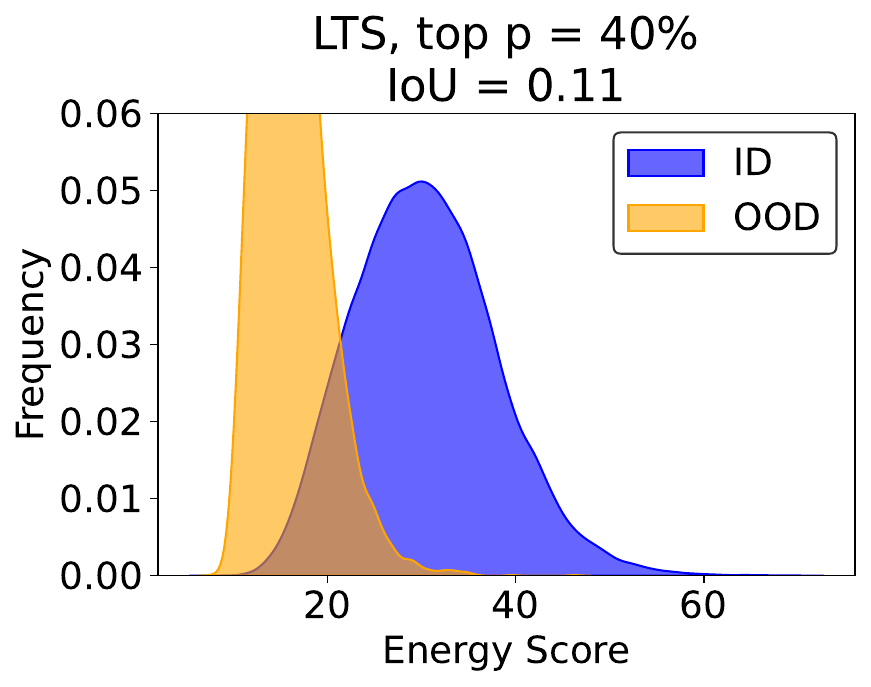}
\includegraphics[width=.3\textwidth]{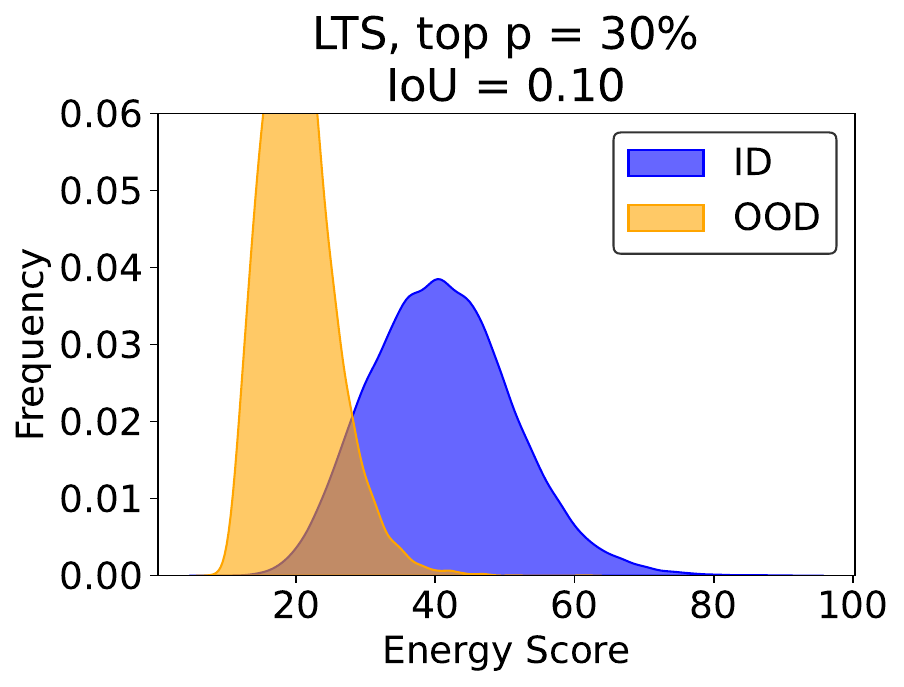}
\includegraphics[width=.3\textwidth]{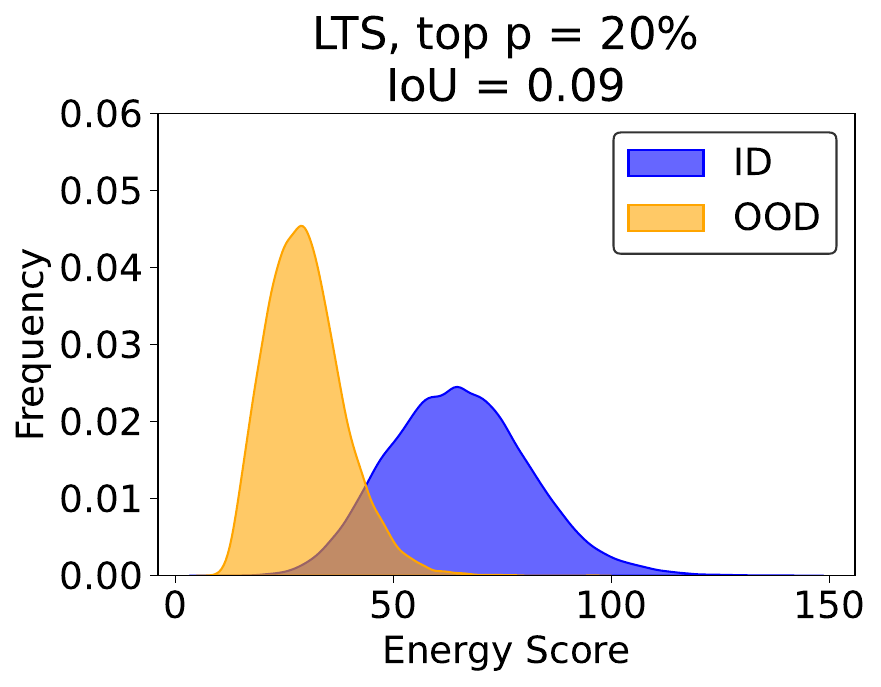}
\includegraphics[width=.3\textwidth]{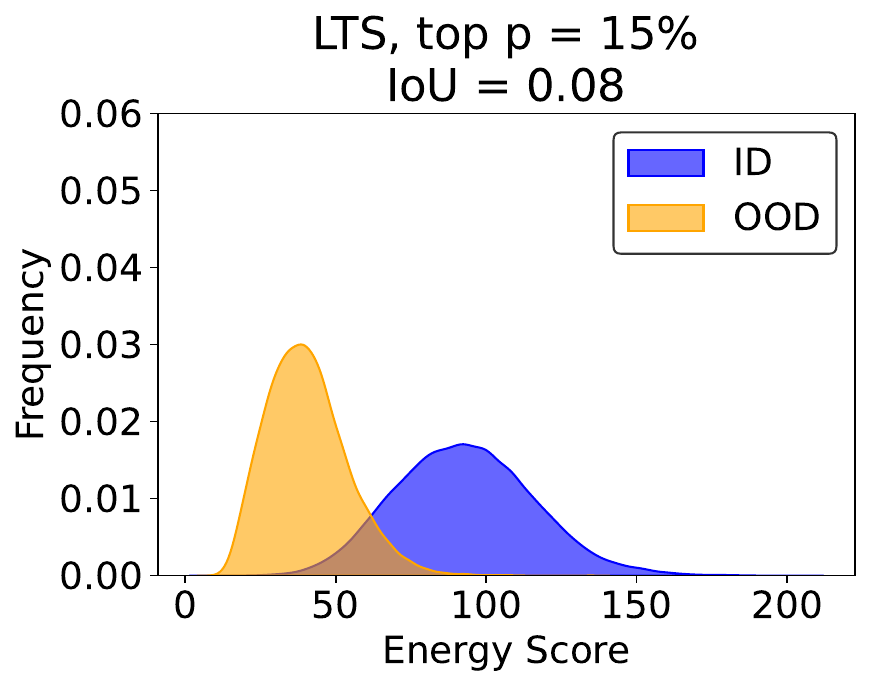}
\includegraphics[width=.3\textwidth]{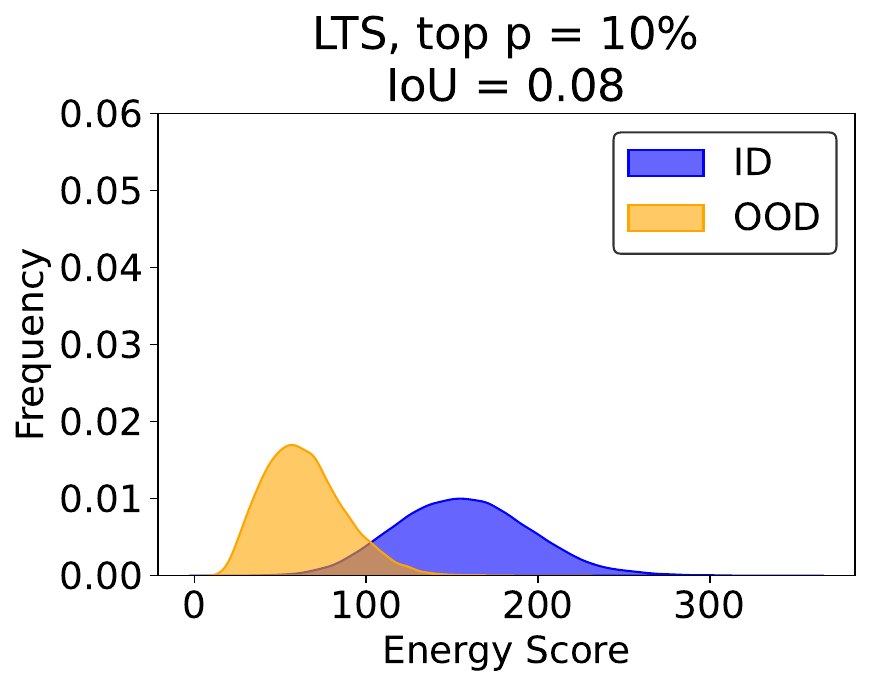}
\includegraphics[width=.3\textwidth]{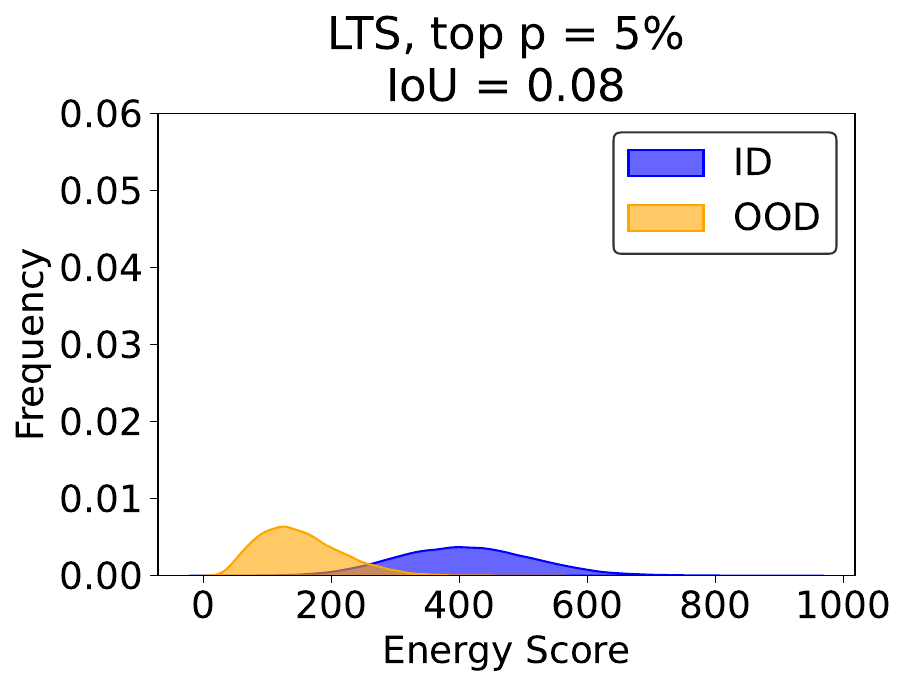}

\caption{\textbf{Energy score distribution morphing as hyperparameter $p$ varies.} Each plot shows how the energy distributions of ID and OOD shift as we apply LTS using the top $p$-th percentile of activation values for logit scaling. The first plot shows the raw energy scores without any LTS adjustment. Progress toward separation is measured by the intersection over union (IoU). The ID dataset is ImageNet-1k; the OOD dataset is iNaturalist. Architecture: ResNet-50.}
\figlabel{morphing}
\end{figure*}

\section{Visualization of failure cases}
\label{appendix_lts_failure_cases}
\begin{figure*}[h]
    \centering
    \begin{minipage}{\textwidth}
        \centering
        ImageNet-1k (ID) samples misclassified as OOD\\[0.3em]
        \includegraphics[width=0.99\textwidth]{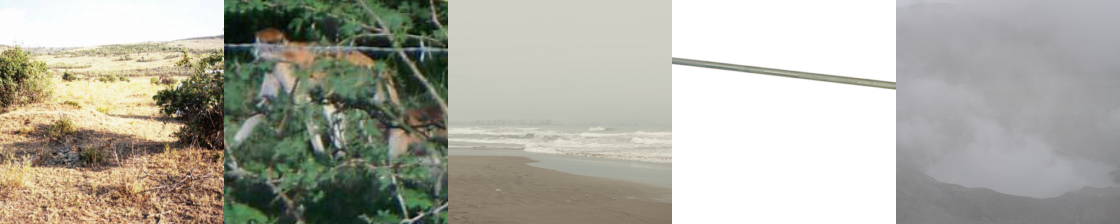}
    \end{minipage}
    
    \vspace{0.5em}
    
    \begin{minipage}{\textwidth}
        \centering
        iNaturalist samples misclassified as ID\\[0.3em]
        \includegraphics[width=0.99\textwidth]{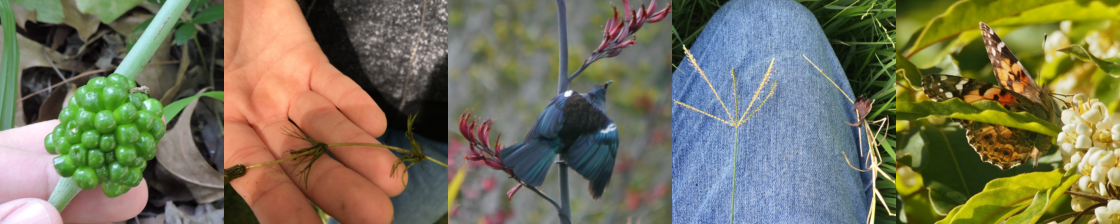}
    \end{minipage}
    \vspace{0.5em}
    
    \begin{minipage}{\textwidth}
        \centering
        Ninco samples misclassified as ID\\[0.3em]
        \includegraphics[width=0.99\textwidth]{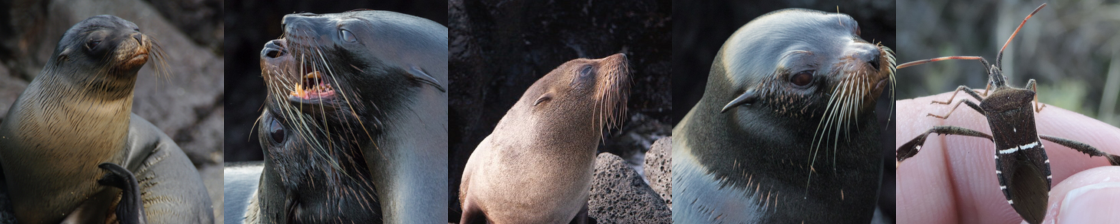}
    \end{minipage}
    \vspace{0.5em}
    
    \begin{minipage}{\textwidth}
        \centering
        Texture samples misclassified as ID\\[0.3em]
        \includegraphics[width=0.99\textwidth]{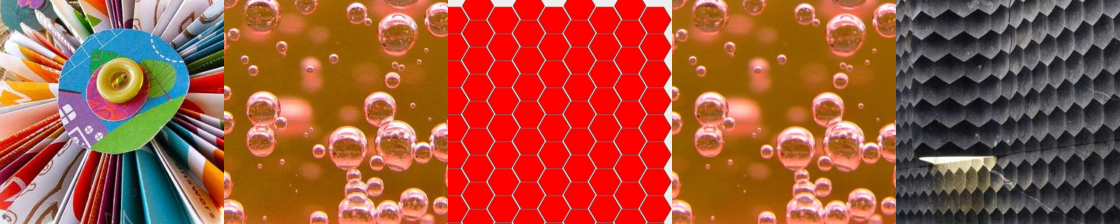}
    \end{minipage}
    \vspace{0.5em}
    
    \begin{minipage}{\textwidth}
        \centering
        OpenImage-O samples misclassified as ID\\[0.3em]
        \includegraphics[width=0.99\textwidth]{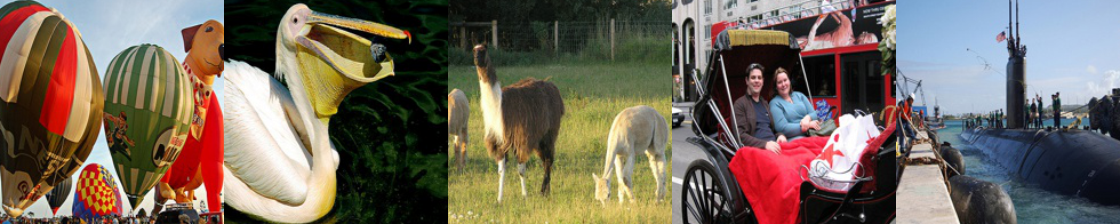}
    \end{minipage}
    \vspace{1em}
    
    \caption{\textbf{Examples of failure cases of LTS.} Top row: ImageNet-1k (ID) samples misclassified as OOD, often due to low-texture scenes or atypical visual features. Rows 2–5: OOD samples from iNaturalist, Ninco, Texture, and OpenImage-O datasets misclassified as ID, typically because they exhibit strong visual or semantic similarity to natural ImageNet classes, or because their domain shift is relatively subtle. These examples highlight the main challenges faced by LTS in distinguishing between ID and OOD samples, particularly under texture-biased scenarios. Architecture used for these results is ResNet-50.}
    \label{fig:failure_cases}
\end{figure*}

\begin{figure*}[h]
    \centering
    \begin{minipage}{\textwidth}
        \centering
        ImageNet-1k (ID) samples misclassified as OOD\\[0.3em]
        \includegraphics[width=0.99\textwidth]{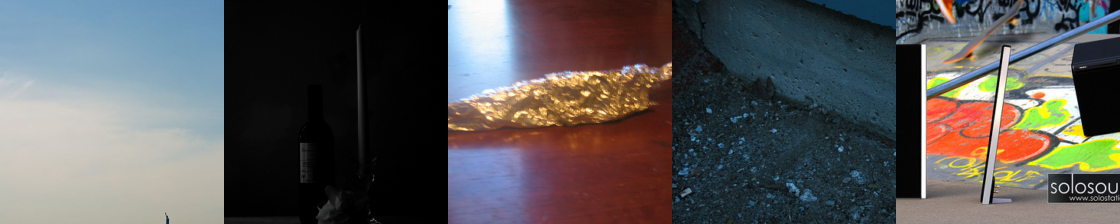}
    \end{minipage}
    
    \vspace{0.5em}
    
    \begin{minipage}{\textwidth}
        \centering
        iNaturalist samples misclassified as ID\\[0.3em]
        \includegraphics[width=0.99\textwidth]{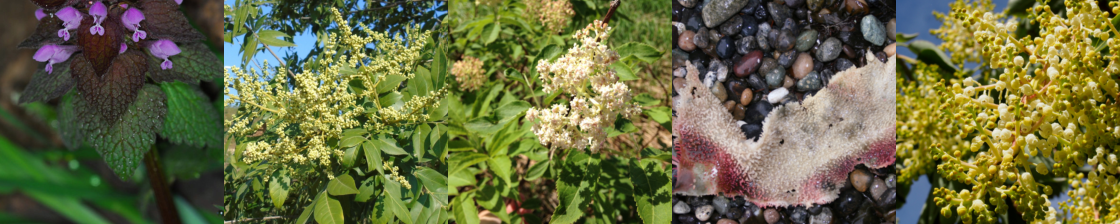}
    \end{minipage}
    \vspace{0.5em}
    
    \begin{minipage}{\textwidth}
        \centering
        Sun samples misclassified as ID\\[0.3em]
        \includegraphics[width=0.99\textwidth]{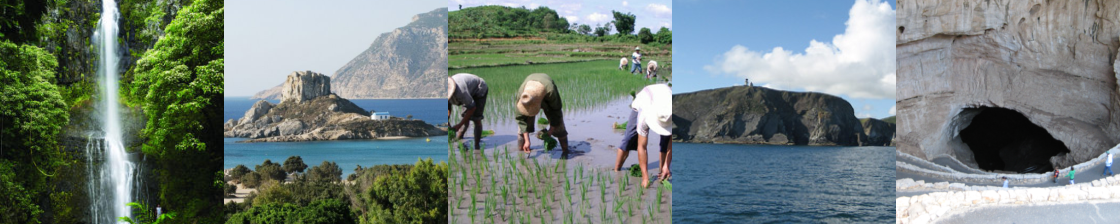}
    \end{minipage}
    \vspace{0.5em}
    
    \begin{minipage}{\textwidth}
        \centering
        Texture samples misclassified as ID\\[0.3em]
        \includegraphics[width=0.99\textwidth]{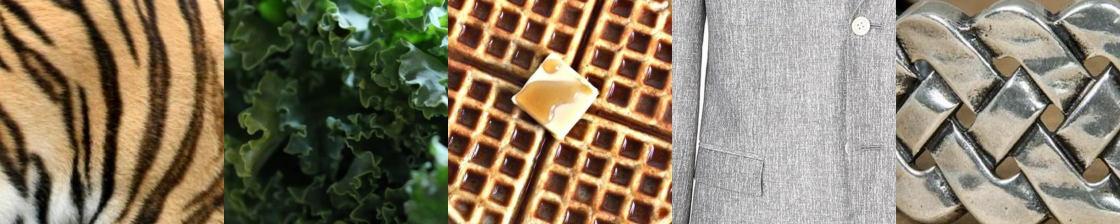}
    \end{minipage}
    \vspace{0.5em}
    
    \begin{minipage}{\textwidth}
        \centering
        Places samples misclassified as ID\\[0.3em]
        \includegraphics[width=0.99\textwidth]{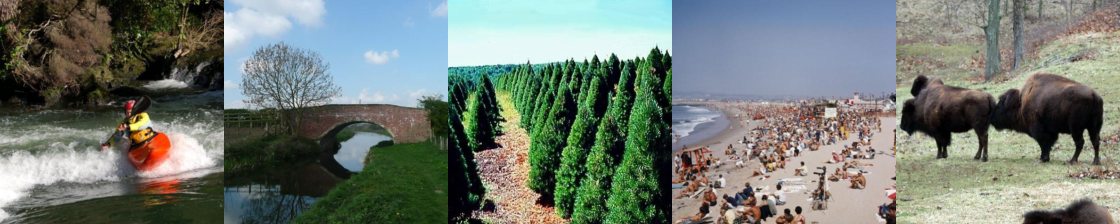}
    \end{minipage}
    \vspace{1em}
    
    \caption{\textbf{Illustration of common LTS failure cases on the ViT-B-16 architecture.} ImageNet-1k samples are misclassified as OOD (top row) mostly due to images with simple and plain backgrounds, while OOD samples from iNaturalist, SUN, Textures, and Places are misclassified as ID (rows 2-5) mostly due to visual similarity with ID data, showing limitations of LTS on ViT-B-16 architecture.}
    \label{fig:failure_cases_vit}
\end{figure*}

\begin{figure*}[h]
    \centering
    \begin{minipage}{\textwidth}
        \centering
        ImageNet-1k (ID) samples misclassified as OOD\\[0.3em]
        \includegraphics[width=0.99\textwidth]{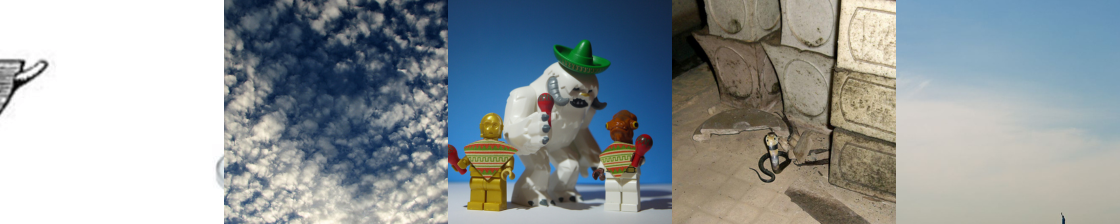}
    \end{minipage}
    
    \vspace{0.5em}
    
    \begin{minipage}{\textwidth}
        \centering
        iNaturalist samples misclassified as ID\\[0.3em]
        \includegraphics[width=0.99\textwidth]{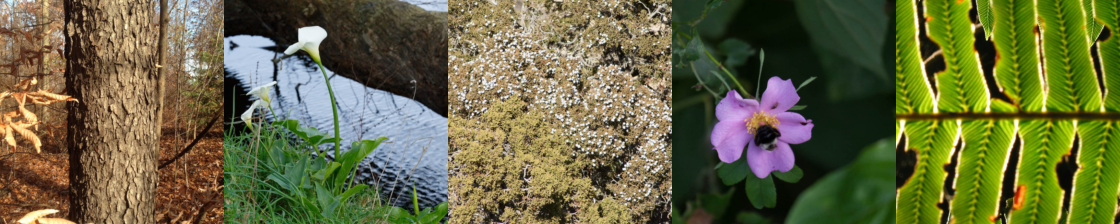}
    \end{minipage}
    \vspace{0.5em}
    
    \begin{minipage}{\textwidth}
        \centering
        Sun samples misclassified as ID\\[0.3em]
        \includegraphics[width=0.99\textwidth]{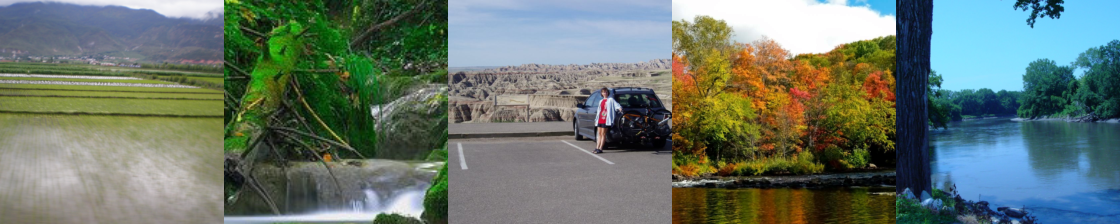}
    \end{minipage}
    \vspace{0.5em}
    
    \begin{minipage}{\textwidth}
        \centering
        Texture samples misclassified as ID\\[0.3em]
        \includegraphics[width=0.99\textwidth]{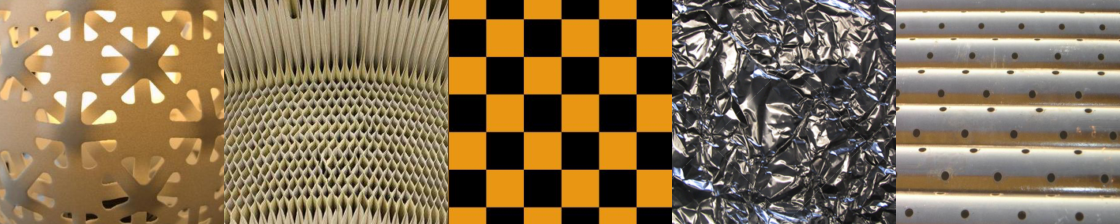}
    \end{minipage}
    \vspace{0.5em}
    
    \begin{minipage}{\textwidth}
        \centering
        Places samples misclassified as ID\\[0.3em]
        \includegraphics[width=0.99\textwidth]{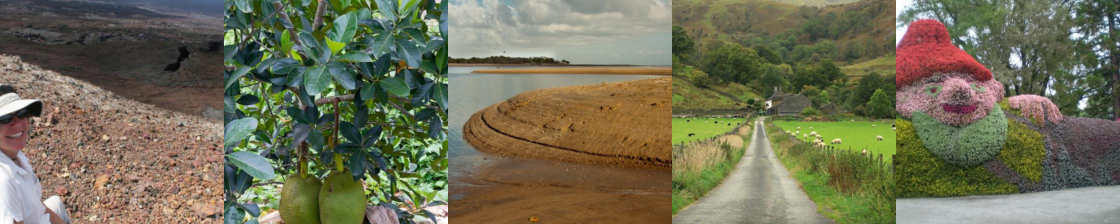}
    \end{minipage}
    \vspace{1em}
    
    \caption{\textbf{Illustration of common LTS failure cases on the Swin-B architecture.}The first row shows mostly clean, varied ImageNet-1k ID samples misclassified as OOD, likely due to minimal context or abstract visual features. Remaining rows include OOD samples from iNaturalist (e.g., tree bark, flowers, and textures), SUN (natural landscapes), Textures (highly regular patterns), and Places (natural and manmade scenes). These OOD images often resemble ID data in structure, color, or composition, leading the method to mistakenly classify them as ID.} 
    \label{fig:failure_cases_swin}
\end{figure*}

We provide a visualization of the most common failure cases in \figref{failure_cases}. Specifically, we show examples where ImageNet-1k (ID) samples are misclassified as OOD, and where OOD samples from iNaturalist, Ninco, Texture, and OpenImage-O are incorrectly classified as ID on ResNet-50 architecture. Additionally, we present misclassified examples for ViT-B (\figref{failure_cases_vit}) and Swin-B (\figref{failure_cases_swin}) architectures, where ImageNet-1k serves as the ID dataset and iNaturalist, Places, SUN, and Texture are OOD datasets. These cases illustrate typical challenges for the model, including subtle domain shifts, texture bias, and high semantic similarity between ID and OOD samples.

\clearpage
\clearpage

\bibliography{refs}

\end{document}